\documentclass[lettersize,journal]{IEEEtran}
\usepackage{amsmath,amsfonts}
\usepackage{algorithmic}
\usepackage{algorithm}
\usepackage[dvipsnames, table]{xcolor}
\usepackage{array}
\usepackage[caption=false,font=footnotesize,labelfont=rm,textfont=rm]{subfig}
\usepackage{textcomp}
\usepackage{stfloats}
\usepackage{url}
\usepackage{bbding}
\usepackage{color}
\usepackage{verbatim}
\usepackage{hyperref}
\usepackage{colortbl}
\definecolor{colorh}{rgb}{1,0.60,0.20}
\definecolor{colorm}{rgb}{1,0.72,0.30}
\definecolor{colorl}{rgb}{1,0.88,0.70}

\usepackage{orcidlink}
\usepackage{graphicx}
\usepackage{cite}
\usepackage{microtype}
\usepackage{graphicx}
\usepackage{subfig}
\usepackage{graphicx}
\usepackage{float}
\usepackage{multirow}
\usepackage{bm}
\usepackage{booktabs}
\usepackage{amsmath}
\usepackage{amssymb}
\usepackage{mathtools}
\usepackage{amsthm}

\begin{document}

\title{LLIC: Large Receptive Field Transform Coding with Adaptive Weights for Learned Image Compression}

\author{Wei Jiang~\orcidlink{0000-0001-9169-1924}, Peirong Ning~\orcidlink{0009-0006-6645-5130}, Jiayu Yang~\orcidlink{0000-0001-9729-1294}, Yongqi Zhai~\orcidlink{0000-0002-3748-1392}, Feng Gao~\orcidlink{0009-0006-1843-3180},~\IEEEmembership{Member,~IEEE},\\Ronggang Wang~\orcidlink{0000-0003-0873-0465},~\IEEEmembership{Member,~IEEE}
\thanks{
This work is financially supported for Outstanding Talents Training Fund in Shenzhen, Shenzhen Science and Technology Program-Shenzhen Cultivation of Excellent Scientific and Technological Innovation Talents project (Grant No. RCJC20200714114435057) , Shenzhen Science and Technology Program-Shenzhen Hong Kong joint funding project (Grant No. SGDX20211123144400001), National Natural Science Foundation of China U21B2012, R24115SG MIGU-PKU META VISION TECHNOLOGY INNOVATION LAB.
}
\thanks{Wei Jiang, Peirong Ning, Jiayu Yang, Yongqi Zhai, Ronggang Wang are
with School of Electronic and Computer Engineering, Peking University, 518055 Shenzhen, China (email: \url{wei.jiang1999@outlook.com}).}
\thanks{Jiayu Yang, Yongqi Zhai, Ronggang Wang are with Peng Cheng Laboratory, 518000
Shenzhen, China (email: \url{rgwang@pkusz.edu.cn}).}
\thanks{Feng Gao is with School of Arts, Peking University, 100871, Beijing, China.}
\thanks{Ronggang Wang is the corresponding author.}
\thanks{Digital Object Identifier 10.1109/TMM.2024.3416831}
}

\markboth{
  IEEE Transactions on Multimedia}%
{Shell \MakeLowercase{\textit{et al.}}: A Sample Article Using IEEEtran.cls for IEEE Journals}


\maketitle

\begin{abstract}
    The effective receptive field (ERF) plays an important role in transform coding, which determines how much redundancy can be removed during
    transform and how many spatial priors can be utilized to synthesize textures during inverse transform.
    Existing methods rely on stacks of small kernels, whose {ERFs remain} {insufficiently large}, or heavy non-local attention mechanisms,
    which limit the potential of high-resolution image coding. To tackle this issue, we propose Large Receptive Field Transform Coding with
    Adaptive Weights for Learned Image Compression (LLIC). Specifically, for the \textit{first} time in the learned image
    compression community, we introduce \textit{a few} large kernel-based depth-wise convolutions to reduce more redundancy while maintaining modest complexity.
    Due to the wide range of image diversity, {we further propose a mechanism to augment convolution adaptability through the self-conditioned generation of weights.}
    The large kernels cooperate with non-linear embedding and gate mechanisms for better expressiveness and 
    lighter point-wise interactions. 
    {Our investigation extends to refined training methods that unlock the full potential of these large kernels.
    Moreover, to promote more dynamic inter-channel interactions, we introduce an adaptive channel-wise bit allocation strategy that autonomously generates channel importance factors in a self-conditioned manner.}
    To demonstrate the effectiveness of the proposed transform coding, we align the entropy model
    to compare with existing transform methods and obtain models LLIC-STF, LLIC-ELIC, {and} LLIC-TCM.
    {Extensive experiments demonstrate {that} our proposed LLIC models have significant improvements over {the} corresponding 
    baselines and reduce {the} BD-Rate by $9.49\%, 9.47\%, 10.94\%$ on Kodak over VTM-17.0 Intra, respectively.}
    Our LLIC models achieve state-of-the-art performances and better {trade-offs} between performance and complexity.
\end{abstract}


\section{Introduction}\label{sec:intro}
Learned image compression~\cite{balle2016end,theis2017lossy,balle2018variational,minnen2018joint,minnen2020channel,ma2019image} {has become} an active research area in recent years.
Several models~\cite{zhu2022transformerbased,zou2022the,zhu2022unified,koyuncu2022contextformer,feng2023nvtc,he2022elic,jiang2022mlic,jiang2023mlic++,liu2023learned}
have surpassed the advanced non-learned image codec Versatile Video Coding (VVC) Intra~\cite{bross2021vvc}. 
Most learned image compression models are based on variational autoencoders (VAEs)~\cite{kingma2014vae}.
In this framework, {an analysis transform first converts the input image to a latent representation,
which is then quantized for entropy coding.
A synthesis transform subsequently maps the quantized latent representation back to pixels}.
The advantages of learned image compression over non-learned codecs are 
the end-to-end optimization and \textit{non-linear} transform coding~\cite{balle2020nonlinear}.
\par
{Recent improvements in the rate-distortion performance of learned image compression
can be attributed to advancements in nonlinear transform coding~\cite{balle2020nonlinear}}.
The inherent non-linearity of such coding empowers the conversion of an input image to a more compact latent representation,
typically requiring fewer bits for compression.
\begin{figure}[t]
    \centering
    \includegraphics[width=\linewidth]
    {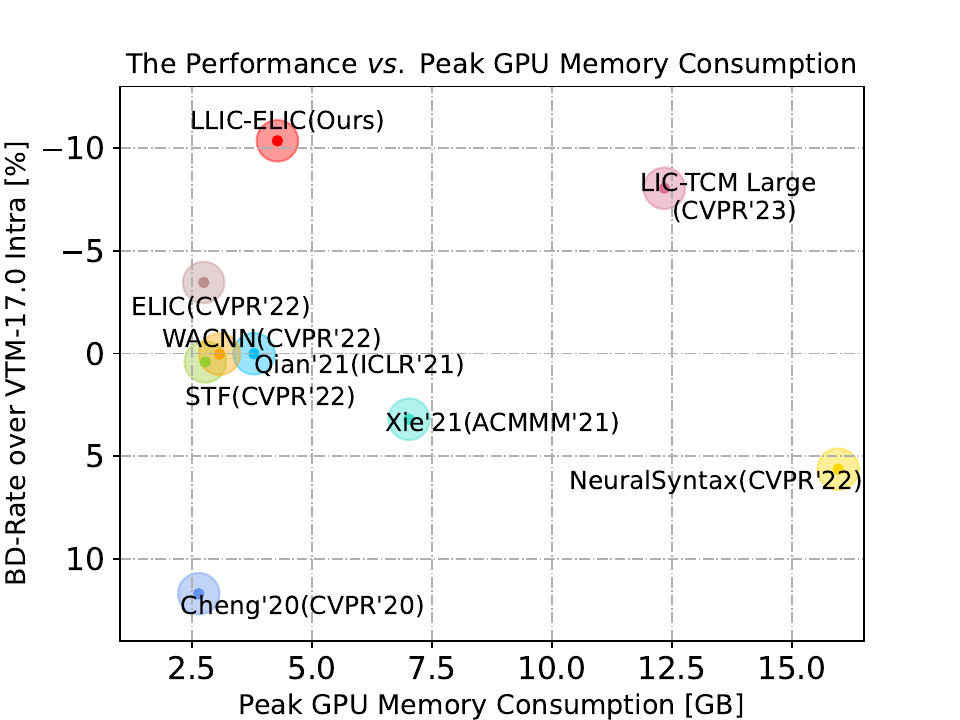}
    \caption{BD-Rate-peak GPU Memory Consumption during testing on CLIC Pro Valid~\cite{CLIC2020} with 2K resolution.
    Our LLIC-ELIC achieves a better trade-off between performance and GPU memory consumption.
    }
    \label{fig:ctx_compare_new}
  \end{figure}
  {The receptive field plays a crucial role in transform coding {because} it determines
  the maximum amount of redundancy that can be removed and
  the compactness of the latent representation. 
  Furthermore, a large receptive field facilitates the generation of more accurate textures from more spatial priors during the synthesis transform.}
  The effective receptive field is {particularly} important for high-resolution image coding
  due to the presence of more spatial correlations. For instance,
in VVC, the largest size of coding unit is $128\times 128$
($64\times 64$ in High Efficiency Video Coding),
{ensuring that long-range dependencies can be captured in high-resolution images.}
Most earlier nonlinear transform coding techniques~\cite{balle2016end, balle2018variational,minnen2018joint,minnen2020channel,cheng2020learned,he2022elic,zhu2022unified,feng2023nvtc,jiang2022mlic,jiang2023mlic++,chen2022two,xie2021enhanced,wang2022neural}
rely on stacks of small kernels (e.g., $3\times 3, 5\times 5$)
to enlarge the receptive field. 
However, the effective receptive fields (ERFs)~\cite{luo2016understanding} of stacks of small kernels {remain} limited,
{as illustrated} in Fig.~\ref{fig:erf}.
To enlarge the receptive field, non-local attention~\cite{zhang2018residual,chen2021nic,liu2023learned}
{has been} employed; however, its inherent quadratic complexity limits the potential for high-resolution image coding.
Overall, 
enlarging the effective receptive field of nonlinear transform coding with acceptable complexity remains a challenge.\par
To our knowledge, large kernel design {with the aim of achieving} a large receptive field
for transform coding has been overlooked.
{Could the application of a few larger kernels potentially confers advantages over the conventional approach of integrating numerous smaller kernels in transform coding?}
{Additionally, due to the wide range of image diversity, the optimal transform coding for each image may differ.}
Compared with existing attention-based or transformer-based transform coding
techniques~\cite{chen2021nic,cheng2020learned,zou2022the,he2022elic,zhu2022transformerbased,liu2023learned},
the fixed convolutional weights {used} during testing are less flexible and adaptive.
Based on above considerations, we introduce a novel Spatial Transform Block (STB)
with large receptive fields and adaptability. Specifically, for the \textit{first}
time in the learned image compression community,
we propose {the application of} depth-wise large kernels
whose sizes range from $9\times 9$ to $11\times 11$ to achieve
a good trade-off between performance and efficiency.
To enhance the \textit{adaptability} of convolution weights, 
we propose generating depth-wise kernel weights by using the input
feature as the condition with a progressive down-sampling strategy.
To argument the non-linearity of {the} transform, we propose the depth-wise residual block
for non-linear embedding. Moreover, a gate block is proposed
for low-complexity point-wise interactions.
We also propose a simple yet effective training technique {that} uses large patches to fully exploit the potential of large kernels. 
Thanks to the proposed spatial transform with large receptive fields and adaptability, 
our proposed transform coding achieves much larger effective receptive fields with smaller depth, as illustrated in Fig.~\ref{fig:erf}.
{ A more widely distributed green
area indicates a larger ERF. {A} larger ERF indicates that more spatial contextual information is captured and
utilized during the transform, {thus making} the latent representation more compact,
which {means}  {that} it requires fewer bits to compress an image.}
Moreover,
the complexity of {the} proposed transform coding method is still modest.
\par
Large adaptive depth-wise kernels are effective {at reducing} spatial redundancy; however, the 
adaptive interactions among channels are limited.
{Given the heterogeneity of information carried by different channels in a latent representation, 
a more targeted approach that dynamically allocates more bits to more informative channels can enhance coding efficiency.}
To address this issue, inspired by Channel Attention~\cite{hu2018squeeze}, we propose a novel adaptive channel transform block (CTB).
The CTB shares the macro architecture of the STB, where non-linear embedding and gate block are also employed.
{Specifically, it generates channel importance factors through a condition-based, progressive down-sampling process, 
subsequently applying these factors to the latent representation to modulate channel-wise importance dynamically.}\par
To validate the effectiveness of {the} proposed transform coding, 
{we combine the proposed transform coding 
techniques with the entropy models of STF~\cite{zou2022the}, ELIC~\cite{he2022elic}, and LIC-TCM~\cite{liu2023learned}}
and obtain the learned image compression models LLIC-STF, LLIC-ELIC, and LLIC-TCM for fair comparisons with existing 
state-of-the-art transform coding methods.
Extensive experiments demonstrate that our models significantly {improve upon} the corresponding baselines,
especially on high-resolution images.
Our LLIC models achieve state-of-the-art performance regarding rate-distortion performance and model complexity (Fig.~\ref{fig:ctx_compare_new},~\ref{fig:complex}, Table~\ref{tab:rd}).
\par
Our contributions are as follows:
\begin{itemize}
    \item
    We introduce the Spatial Transform Block,
    utilizing $11\times 11$ and $9\times 9$ large receptive field transform{s}
    to reduce spatial redundancy.
    To our knowledge, this is the \textit{first} time that {large kernels have been employed} in the learned image compression community.
    \item 
    We propose the generation of depthwise convolutional
    weights in a progressive down-sampling manner, using the input itself as a condition, {making} CNNs adaptive.
    \item
    We propose the Channel Transform Block,
    which employs importance factors for self-conditioned adaptive channel-wise bit allocation in a progressive down-sampling manner,
    using the input itself as a condition.
    \item
    We propose the use of the Depth-wise Residual
    Bottleneck to enhance the non-linearity and the Gate block for efficient point-wise interaction.
    \item
   Extensive experiments demonstrate that our proposed LLIC-STF, LLIC-ELIC, {and} LLIC-TCM have significant improvements
    over corresponding baselines and {reduce BD-Rate by $9.49\%, 9.47\%$, {and} $10.94\%$ on Kodak over VTM-17.0 Intra, respectively}. Our LLIC models achieve state-of-the-art performances and better trade-off between performance and complexity.
\end{itemize}

\section{Related Works}\label{sec:related}
\subsection{Learned Image Compression}\label{sec:related:lic}
Toderici \textit{et al.}~\cite{toderici2015varible,toderici2017full} propose 
the first learned image compression model based on recurrent neural networks,
which encodes the residual between the reconstruction and ground-truth iteratively.
Currently, most of the end-to-end learned image compression models
~\cite{balle2016end,balle2018variational,theis2017lossy,minnen2018joint,minnen2020channel,he2022elic,jiang2022mlic,cheng2020learned}
are based on auto-encoders,
where the input image is first transformed to the latent space for entropy coding, and 
the decoded latent is inversely transformed to the RGB color space.
To enhance the {rate-distortion} performance, Ball{\'e} \textit{et al.}~\cite{balle2018variational} introduce a hyper-prior
for more accurate entropy coding.
Context modeling~\cite{minnen2018joint} is also utilized to explore the correlations
between current symbols and decoded symbols.
Minnen \textit{et al.}~\cite{minnen2018joint} {employ} pixel-cnn~\cite{van2016conditional}
for serial context prediction. He \textit{et al.}~\cite{he2021checkerboard} propose checkerboard context partition
for parallel context modeling.
Minnen \textit{et al.}~\cite{minnen2020channel} propose conducting serial
context modeling along the channel dimension for faster decoding.
In addition, global context modeling~\cite{guo2021causal,jiang2022mlic,jiang2023mlic++} is also introduced to explore the correlations among
distant symbols.
Multi-dimensional context modeling is {recently} developed.
He \textit{el al.}~\cite{he2022elic} {combine} the checkerboard partition and channel-wise context modeling.
Jiang \textit{et al.}~\cite{jiang2022mlic,jiang2023mlic++} propose the multi-reference 
entropy modeling to capture the local, global, and channel-wise correlations in an
entropy model.
\subsection{Learned Transform Coding}\label{sec:related:tc}
The advantage of end-to-end learned image compression over non-learned codecs is its
non-linear transform coding~\cite{balle2020nonlinear}.
{The} transform plays an important role in improving the rate-distortion performance.
For enhanced non-linearity, techniques such as generalized divisive normalization (GDN)~\cite{balle2015gdn} and Residual
Bottleneck~\cite{he2016resnet} have been utilized in learned image compression.
Moreover, innovative normalization methods~\cite{shin2022expanded} have been proposed in recent years.
Furthermore, Cheng \textit{et al.}\cite{cheng2020learned} suggest employing pixel shuffle for improved up-sampling.
{Various approaches}~\cite{akbari2021learned,gao2021neural,chen2022two} {have been} proposed for better interactions between
high-frequency and low-frequency features. For example,
Akbari \textit{et al.}~\cite{akbari2021learned} propose {employing} octave convolution~\cite{chen2019drop}
to preserve more spatial structure of the information.
Gao~\textit{et al.}~\cite{gao2021neural} decompose the images into several layers with different frequency
attributes for greater adaptability.
Xie \textit{et al.}~\cite{xie2021enhanced} propose {the adoption of} invertible neural networks to reduce information loss during transform.
Inspired by non-learned codecs, wavelet-like transform~\cite{ma2019iwave} is also introduced in recent years.
{In addition}, non-local attention~\cite{chen2021nic}, simplified attention~\cite{cheng2020learned},
and group-separated attention~\cite{guo2021causal} are employed to reduce more spatial redundancy. However,
such attention mechanisms are much heavier and lead to {greater} complexity.
Transformers~\cite{vas2017attention,liu2021swin} have been employed in several works~\cite{zhu2022transformerbased,zou2022the,liu2023learned,bai2022towards,jeny2022efficient,lu2021tic,wang2022end}.
{For example, Zou \textit{et al.}~\cite{zou2022the}, Lu \textit{et al.}~\cite{lu2021tic}, Zhu \textit{et al.}~\cite{zhu2022transformerbased}, and Wang \textit{et al.}~\cite{wang2022end}
stack swin-transformer~\cite{liu2021swin} layers in transforms to reduce more redundancy. 
Liu \textit{et al.}~\cite{liu2023learned} employ mixed CNN-Transformer architectures for enhanced 
local and non-local interactions. The dynamic weights and large receptive field of 
transformers contribute significantly to the overall performance enhancement}.\par
However, to our knowledge, the large kernels are still not explored in learned image compression.
Employing large kernels may lead to larger effective receptive fields, which implies
that more redundancy can be reduced. 
Furthermore, how to utilize large kernels without leading to high complexity is \textit{non-trivial}.
\begin{figure*}
    \centering
    \includegraphics[width=0.985\linewidth]
    {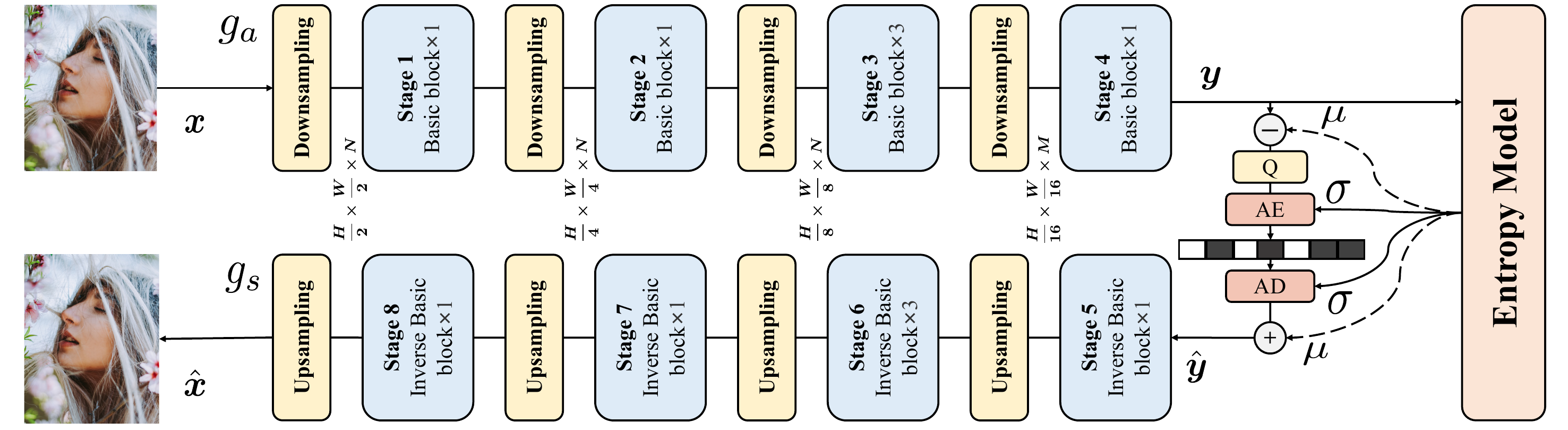}
    \caption{Network architecture of our LLIC-STF, LLIC-ELIC, and LLIC-TCM.
    $g_a$ is the analysis transform. $g_s$ is the synthesis transform. $Q$ is {the} quantization.
    $\mu$ and
    $\sigma$ are the estimated mean and scale of latent $\hat{\boldsymbol{y}}$ for probability estimation.
    Following baseline models, the latent representation $\boldsymbol{y}$
    subtracts the means $\boldsymbol{\mu}$ for quantization before arithmetic encoding (AE), and 
    the decoded residual $Q(\boldsymbol{y} - \boldsymbol{\mu})$ adds the means $\boldsymbol{\mu}$ after arithmetic decoding (AD).
    $N=192, M=320$.}
    \label{fig:arch}
\end{figure*}
\section{Method}\label{sec:method}
\subsection{Problem Formulation}
The formulation of end-to-end optimized image compression is first introduced.
{The fundamental architecture of this paradigm comprises three key components:}
an analysis transform $g_a$, a synthesis transform $g_s$, and an entropy model $p$. The entropy model $p$ computes the means $\boldsymbol{\mu}$ and scales $\boldsymbol{\sigma}$ for probability estimation. 
The input image $\boldsymbol{x}$ is first transformed to the latent representation $\boldsymbol{y}$
via analysis transform $g_a$. {The} latent representation is quantized to $\hat{\boldsymbol{y}}$ for
entropy coding. $\hat{\boldsymbol{y}}$ is inverse transformed to the reconstructed image $\hat{\boldsymbol{x}}$
via the synthesis transform $g_s$. 
{To ensure the differentiability of the model during the training phase, the quantization operation is {replaced}
by either the addition of uniform noise (AUN)~\cite{balle2018variational}
or the straight-through estimator (STE)~\cite{theis2017lossy}.}
{Notably, adding uniform noise $\boldsymbol{u} \sim \mathcal{U}(-0.5, 0.5)$~\cite{balle2018variational}
    to latent representation makes {optimization of} the compression model for rate-distortion performance equivalent to the minimization of the KL divergence (in VAEs~\cite{kingma2014vae})}:
\begin{equation}
\begin{aligned}
    \mathcal{L} &= -(\mathcal{R} + \lambda\times \mathcal{D})\\
    &= \mathbb{E}_{q(\tilde{\boldsymbol{y}}|\boldsymbol{x})}\Biggl[ \underbrace{\log p(\boldsymbol{x}|\tilde{\boldsymbol{y}})}_{\textrm{distortion}} + \underbrace{\log p(\tilde{\boldsymbol{y}})}_{\textrm{rate}} - \underbrace{\log q(\tilde{\boldsymbol{y}}|\boldsymbol{x}}_{0})\Biggr].
\end{aligned}
\end{equation}
$\log p(\boldsymbol{x}|\tilde{\boldsymbol{y}})$ is considered {to be} distortion because
\begin{equation}
    p(\boldsymbol{x}|\tilde{\boldsymbol{y}}) = \mathcal{N}(\boldsymbol{x}|\tilde{\boldsymbol{x}}, (2\lambda)^{-1}\boldsymbol{1}),
\end{equation}
which is the mean square error (MSE). 
$\mathbb{E}_{q(\tilde{\boldsymbol{y}}|\boldsymbol{x})}\log p(\tilde{\boldsymbol{y}})$ is the cross entropy, which is the theoretical bound of entropy coding.
\begin{equation}
    q(\tilde{\boldsymbol{y}}|\boldsymbol{x}) = q(\tilde{\boldsymbol{y}}|\boldsymbol{y}) = \mathcal{U}(\tilde{\boldsymbol{y}}|\boldsymbol{y}-0.5, \boldsymbol{y}+0.5)=1.
\end{equation}
{Currently}, most of the learned image compression models~\cite{balle2018variational,minnen2018joint,cheng2020learned,minnen2020channel,xie2021enhanced,zou2022the,he2022elic,jiang2022mlic}
also incorporate hyper-priors~\cite{balle2018variational}. 
The side information $\boldsymbol{z}$ is extracted from $\boldsymbol{y}$ using a hyper-prior network.
The side information $\tilde{\boldsymbol{z}}$ or $\hat{\boldsymbol{z}}$ helps {to} estimate
the entropy of latent $\tilde{\boldsymbol{y}}$ or $\hat{\boldsymbol{y}}$.
The loss function during training becomes
\begin{equation}
    \label{eq:loss}
    \begin{aligned}
        \mathcal{L} &= -(\mathcal{R} + \lambda\times \mathcal{D})\\
    &= \mathbb{E}_{q(\tilde{\boldsymbol{y}}, \tilde{\boldsymbol{z}}|\boldsymbol{x})}\Biggl[ \underbrace{\log p(\boldsymbol{x}|\tilde{\boldsymbol{y}})}_{\textrm{distortion}} + \underbrace{\log p(\tilde{\boldsymbol{y}}|\tilde{\boldsymbol{z}}) + \log p(\tilde{\boldsymbol{z}})}_{\textrm{rate}} \\&- \underbrace{\log q(\tilde{\boldsymbol{y}}|\boldsymbol{x}) + \log q(\tilde{\boldsymbol{z}}|\boldsymbol{y}}_{0})\Biggr].
    \end{aligned}
\end{equation}\par
Compared with traditional image codecs~\cite{pennebaker1992jpeg,rabbani2002jpeg2000,sullivan2012overview, bross2021vvc}, 
the advantage of the learned image compression is its \textit{non-linear} transform coding~\cite{balle2020nonlinear}. 
In such a framework, the analysis transform and synthesis transform play {important roles}.\par

{First, the transform is employed to de-correlate the input image,
effectively reducing correlations and thus enabling a more compact latent representation. 
This, in turn, contributes to a reduced bit-rate necessary for compression. Despite these advancements, 
the receptive fields of most current learned image compression models remain somewhat limited, 
leading to the persistence of certain redundancies within the latent representation. 
Moreover, the inflexibility of transform module weights during inference, due to their fixed nature, 
significantly impairs content adaptability, thereby constraining overall performance.}\par
Second, from 
the perspective of generative models, {the synthesis transform is conceptualized
as a generator that plays a pivotal role in influencing the quality of the reconstructed image. 
A powerful and expressive synthesis transform has the capacity to produce
finer details from the input latent representation $\hat{\boldsymbol{y}}$, enhancing the visual quality of the output.}\par
Third, most of the analysis transforms down-sample the input four times, which makes
the resolution of the latent representation $\boldsymbol{y}$ much smaller than that of the
input image $\boldsymbol{x}$.
{The complexity of image compression models is primarily attributable to transform coding. 
For instance, as demonstrated by Cheng'20~\cite{cheng2020learned}, the forward MACs (MACs) required
by their model for an input image of size $768\times 512$ reach $415.61$ GMACs,
in stark contrast to the {mere} $0.53$ GMACs necessitated by the entropy model.
Thus, the development of a more potent transform module that can minimize redundancies
while maintaining a reasonable model complexity presents a considerable challenge.}\par
{These} challenges, along with the potential to further enhance the rate-distortion performance 
of learned image compression, motivate us to design large receptive field transform coding with
self-conditioned adaptability for learned image compression.
\begin{figure*}
    \centering
    \includegraphics[width=0.985\linewidth]
    {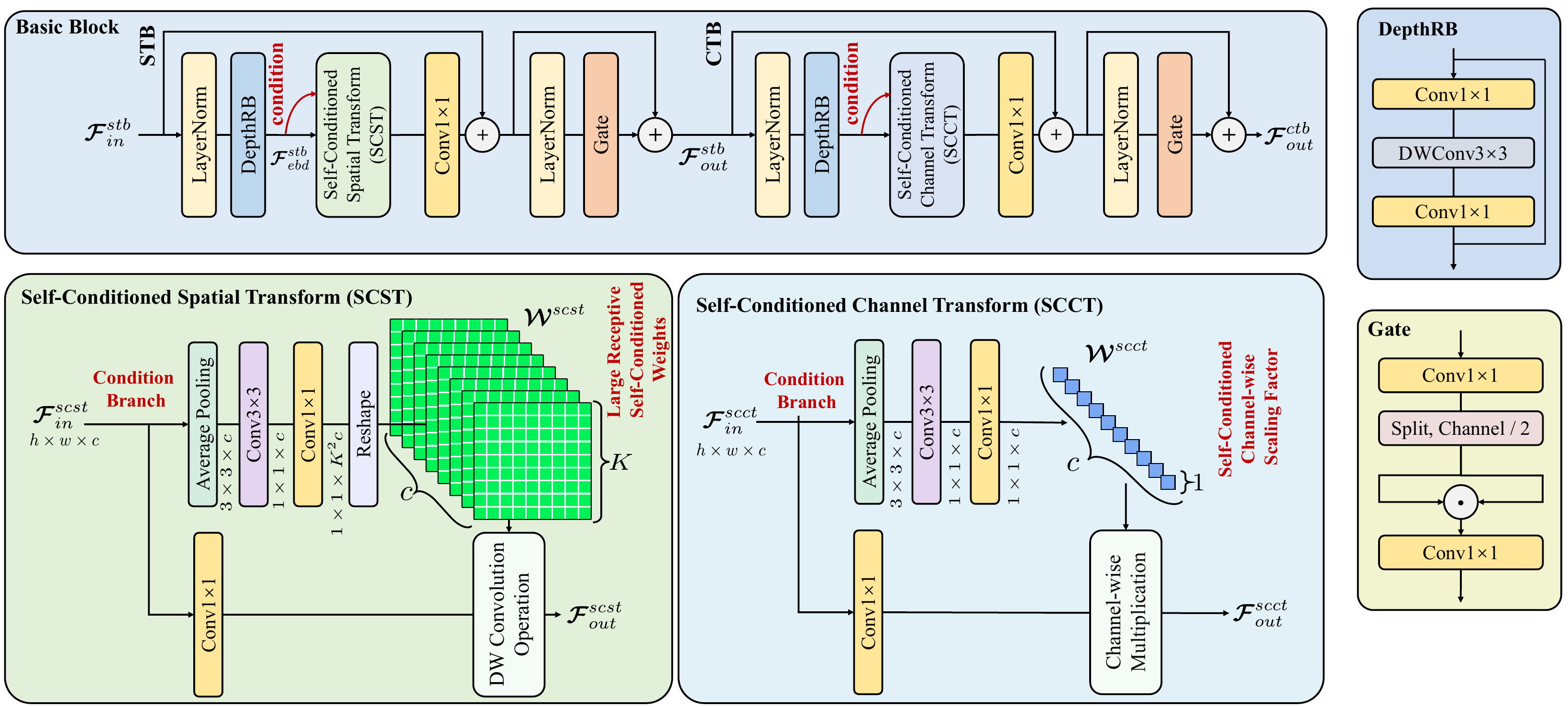}
    \caption{Architecture of the proposed basic block.
    STB is the proposed Spatial Transform Block. CTB is the proposed Channel Transform Block. DepthRB is the depth-wise
    residual block for non-linear embedding. Gate is the proposed Gate Block. $\boldsymbol{\mathcal{F}}_{in}^{stb}$ is the
    input of STB. $\boldsymbol{\mathcal{F}}_{in}^{ctb}$ is the input of CTB.
    {In STB, we employ \textit{large} kernels to capture more spatial contexts, and the kernel size $K$
    is set to $11$ or $9$ in our method}.
    }
    \label{fig:basic}
\end{figure*}
\subsection{Overview of Our Approach}\label{sec:method:overview}
The overall architecture of {the} proposed LLIC-STF, LLIC-ELIC, and LLIC-TCM is
presented in Fig.~\ref{fig:arch} and Fig.~\ref{fig:basic}.
{Our baseline models employ three {types} of state-of-the-art transform coding techniques.}
STF~\cite{zou2022the} employs transformers~\cite{liu2021swin} for more compact latent {representations}.
ELIC~\cite{he2022elic} employs residual convolutional layers and attention techniques~\cite{cheng2020learned}
to enhance the non-linearity.
LIC-TCM~\cite{liu2023learned} employs mixed CNN-transformer architectures to focus 
on the local and non-local redundancies.
For fair comparisons with existing transform coding techniques, {the entropy models of
LLIC-STF, LLIC-ELIC, and LLIC-TCM are aligned with the entropy models} of STF~\cite{zou2022the},
ELIC~\cite{he2022elic}, and LIC-TCM~\cite{liu2023learned}, respectively.\par
{Aligning with our baseline models and established methods~\cite{balle2018variational,minnen2018joint,minnen2020channel,cheng2020learned,jiang2022mlic},
the proposed analysis and synthesis transforms in our work encompass four stages. 
Within each stage, the input undergoes either down-sampling or up-sampling through dedicated blocks.
These down-sampling and up-sampling blocks incorporate a convolutional layer and a depth-wise residual bottleneck~\cite{he2022elic}.
Subsequent to the sampling operations,
the resultant features are propagated through basic blocks or their inverse counterparts.
The basic block comprises two fundamental components: a Spatial Transform Block (STB) and a Channel Transform Block (CTB).
In the inverse basic block, the architectural arrangement is reversed, with the positions of the STB and CTB interchanged.
The structures of STB and CTB follow the classic architecture of Transformers~\cite{vas2017attention,liu2021swin}.
To ensure stable training, both the STB and CTB integrate layer normalization~\cite{ba2016layer}.
Internally, the STB and CTB employ a Depth-wise Residual Block (DepthRB) for non-linear embedding,
coupled with the core transform blocks and a gated block to facilitate point-wise interactions.}
\subsection{Proposed Spatial Transform Block (STB)}\label{sec:method:stb}
The architecture of STB is {shown} in Figure~\ref{fig:basic}.
Depth Residual Block (DepthRB) is proposed for \textit{non-linear} embedding. 
A Gate Block is proposed for efficient point-wise interactions with low complexity.
We propose the Self-Conditioned Spatial Transform (SCST) to effectively reduce spatial
redundancy. The overall process is formulated as:
\begin{equation}
    \begin{aligned}
        \boldsymbol{\mathcal{F}}_{ebd}^{stb} &= \textrm{DepthRB}(\textrm{Norm}(\boldsymbol{\mathcal{F}}_{in}^{stb})),\\
        {\boldsymbol{\mathcal{F}}_{scst}^{stb}} &= \textrm{SCST}(\boldsymbol{\mathcal{F}}_{ebd}^{stb}),\\
        {\boldsymbol{\mathcal{F}}_{skip}^{stb}} &= \textrm{Conv}{1\times1}\left(\boldsymbol{\mathcal{F}}^{scst}\right) +  \boldsymbol{\mathcal{F}}_{in}^{stb},\\
        \boldsymbol{\mathcal{F}}_{out}^{stb} &= \textrm{Gate}(\textrm{Norm}({\boldsymbol{\mathcal{F}}_{skip}^{stb}})) + {\boldsymbol{\mathcal{F}}_{skip}^{stb}},
    \end{aligned}
\end{equation}
where $\boldsymbol{\mathcal{F}}_{in}^{stb}$ is the input feature and $\boldsymbol{\mathcal{F}}_{out}^{stb}$
is the output feature.
\subsubsection{Nonlinear Embedding}
Previous works~\cite{vas2017attention,liu2021swin,zou2022the,liu2023learned} adopt one linear convolutional layer for embedding.
In contrast to previous works, our STB uses a non-linear embedding method.
We propose {employing} DepthRB.
The architecture of DepthRB is presented in Figure~\ref{fig:basic}, which
contains a $1\times 1$ convolutional layer, a $3\times 3$ depth-wise
convolutional layer, and a $1\times 1$ convolutional layer.
The use of depth-wise convolution helps to minimize complexity.
The $3\times 3$ depth-wise convolutional layer helps to aggregate more spatial information,
and non-linearity {enhances} the expressiveness of the network.
\subsubsection{Self-Conditioned Spatial Transform (SCST)}
During the transform, {the elimination of redundancies directly correlates with the compactness of the latent representation obtained.
However, current state-of-the-art learned image compression models often fall short in terms of the 
size of their effective receptive fields, as depicted in Fig.~\ref{fig:erf}.} 
They~\cite{cheng2020learned,he2022elic,minnen2018joint,minnen2020channel} usually employ $3\times 3$ or $5\times 5$ kernels.
{This limitation in receptive field size leaves unaddressed redundancies, thereby hampering the efficiency of compression.}
To achieve larger effective receptive fields,
we propose {employing} large depth-wise kernels for analysis and synthesis transform.
To our knowledge, this is the \textit{first} attempt in the learned image compression community to 
apply large kernels for transform.
In our proposed transform coding method, the kernel size is enlarged to $\{11, 11, 9, 9\}$
for different stages of analysis transform.
The kernel sizes of the first two stages are larger because of the larger resolutions of {the}
input features {in} the first two stages.
The large-resolution features {contain} more spatial redundancies.
\par
Although large kernels are employed in our proposed LLIC, the complexity is still modest due to the 
depth-wise connectivity. It is assumed that the large kernel size of a depth-wise convolution is $K_L$, the small kernel size of a vanilla
convolution is $K_S$, and the input and output channels are $N$. The complexity of the large depth-wise convolution is $K_L^2\times N$, and 
the complexity of the small vanilla convolution is $K_S^2\times N^2$.
\begin{equation}
    \begin{aligned}
        K_L^2\times N &\geq K_S^2\times N^2,\\
        K_L &\geq {K_S}\sqrt{N}.\\
    \end{aligned}
\end{equation}
$N$ is $192$ in our LLIC models and our baselines~\cite{he2022elic,zou2022the,balle2018variational,minnen2018joint,minnen2020channel,cheng2020learned},
and $\sqrt{192}\approx 13.86$, indicating that a $11\times 11$ depth-wise convolution is lighter than
a $1\times 1$ vanilla convolution.\par
{In contrast to the existing methods that rely on transformer-based or attention-based transform coding techniques, 
the static convolutional layer weights present a limitation in harnessing the characteristics of the input image or features.}
To this end,
we propose {generating} convolutional weights {by} treating 
the input itself as the condition in a progressive down-sampling manner. 
The condition branch in SCST is utilized to generate the
self-conditioned adaptive weights. Specifically, 
the input feature {$\boldsymbol{\mathcal{F}}_{in}^{scst}\in \mathbb{R}^{c\times h\times w}$}
is first average pooled to $\boldsymbol{\mathcal{F}}_{pool}^{scst}\in \mathbb{R}^{c\times 3\times 3}$,
where $c,h,w$ are the channel number, height, and width of $\boldsymbol{\mathcal{F}}_{in}^{scst}$, respectively.
The self-conditioned adaptive weights {$\boldsymbol{\mathcal{W}}^{scst} \in \mathbb{R}^{c\times K^2}$} are computed 
via the convolution between the average pooled feature $\boldsymbol{\mathcal{F}}_{pool}^{scst}$ and 
the weights of the condition branch, where
$K$ is the kernel size. \par
The overall process of the proposed SCST is formulated as follows:
\begin{equation}
    \begin{aligned}
        \boldsymbol{\mathcal{F}}_{pool}^{scst} &= \textrm{AvgPool}({\boldsymbol{\mathcal{F}}_{in}^{scst}}),\\
        {\boldsymbol{\mathcal{W}}^{scst}} &= \textrm{Conv}1\times1(\textrm{Conv}3\times3(\boldsymbol{\mathcal{F}}_{pool}^{stb})),\\
        {\boldsymbol{\mathcal{F}}^{scst}_{out}} &= {\boldsymbol{\mathcal{W}}^{scst}}\otimes  \textrm{Conv1}\times1({\boldsymbol{\mathcal{F}}_{in}^{scst}}),\\
    \end{aligned}
\end{equation}
where $\boldsymbol{\mathcal{F}}^{scst}_{out}$ is the output feature, $\otimes$ is the convolution operation.
\subsubsection{Gate Mechanism}
In previous works~\cite{zou2022the,zhu2022transformerbased}, the feed forward network (FFN)~\cite{vas2017attention} is adopted
for channel-wise interactions, which contains two linear layers and
a GELU~\cite{hendrycks2016gaussian} activation function. The overall process of {an} FFN is
\begin{equation}
    \begin{aligned}
        \boldsymbol{\mathcal{F}}_{inc}^{gate} &= \textrm{Conv}1\times1(\boldsymbol{\mathcal{F}}_{in}^{gate}),\\
        \boldsymbol{\mathcal{F}}_{act}^{gate} &= \frac{1}{2}\boldsymbol{\mathcal{F}}_{inc}^{gate}\left(1 + \textrm{tanh}\left[ \sqrt{\frac{2}{\pi}}\left( 0.044715\left(\boldsymbol{\mathcal{F}}_{inc}^{gate}\right)^3 \right) \right]\right), \\
        \boldsymbol{\mathcal{F}}_{out}^{gate} &= \textrm{Conv}1\times1(\boldsymbol{\mathcal{F}}_{act}^{gate}) + \boldsymbol{\mathcal{F}}_{in}^{gate},
    \end{aligned}
\end{equation}
where $\boldsymbol{\mathcal{F}}_{in}^{gate} \in \mathbb{R}^{c\times h \times w}$ is the input feature and $\boldsymbol{\mathcal{F}}_{out}^{gate} \in \mathbb{R}^{c\times h\times w}$
is the output feature.
Following existing methods, the channel number of $\boldsymbol{\mathcal{F}}_{inc}^{gate}$ and $\boldsymbol{\mathcal{F}}_{act}^{gate}$
is $2c$.
The first $1\times 1$ convolutional layer increases the dimension of the input, and the second
$1\times 1$ convolutional layer decreases the dimension to the original dimension.
GELU~\cite{hendrycks2016gaussian} is much more complex than ReLU~\cite{nair2010rectified} or LeakyReLU.
The GELU function can be simplified as $\boldsymbol{\mathcal{F}}_{inc}^{gate}\Phi (\boldsymbol{\mathcal{F}}_{inc}^{gate})$~\cite{chen2022simple},
which is quite similar to the gate mechanism. 
Replacing GELU with a gate mechanism leads to lower complexity,
which inspired us to employ a gate block instead of an FFN.
The architecture of our gate block is illustrated in Figure~\ref{fig:basic}.
Specifically, $\boldsymbol{\mathcal{F}}_{mid}^{gate}$ is split into two features
$\boldsymbol{\mathcal{F}}_{1}^{gate} \in \mathbb{R}^{c\times h\times w}$ and $\boldsymbol{\mathcal{F}}_{2}^{gate}\in \mathbb{R}^{c\times h\times w}$ along the channel dimension.
The activated feature $\boldsymbol{\mathcal{F}}_{act}^{gate}$ is obtained via 
the Hadamard product between $\boldsymbol{\mathcal{F}}_{1}^{gate}$ and $\boldsymbol{\mathcal{F}}_{2}^{gate}$.
The gate mechanism causes non-linearity, which is similar to GELU.
The learnable $1\times 1$ convolutional layer also makes the proposed
gate mechanism more flexible. The overall process of the proposed gate block is
\begin{equation}
    \begin{aligned}
        \boldsymbol{\mathcal{F}}_{inc}^{gate} &= \textrm{Conv}1\times1(\boldsymbol{\mathcal{F}}_{in}^{gate}),\\
        \boldsymbol{\mathcal{F}}_{1}^{gate}, \color{blue}{\boldsymbol{\mathcal{F}}_{2}^{gate}} &= \textrm{Split}(\boldsymbol{\mathcal{F}}_{inc}^{gate}),\\
        \boldsymbol{\mathcal{F}}_{act}^{gate} &= \boldsymbol{\mathcal{F}}_{1}^{gate} \odot \boldsymbol{\mathcal{F}}_{2}^{gate},\\
        \boldsymbol{\mathcal{F}}_{out}^{gate} &= \textrm{Conv}1\times1(\boldsymbol{\mathcal{F}}_{act}^{gate}) + \boldsymbol{\mathcal{F}}_{in}^{gate}.
    \end{aligned}
\end{equation}
\begin{figure*}[t]
	\centering
	\subfloat{
		\includegraphics[scale=0.43]{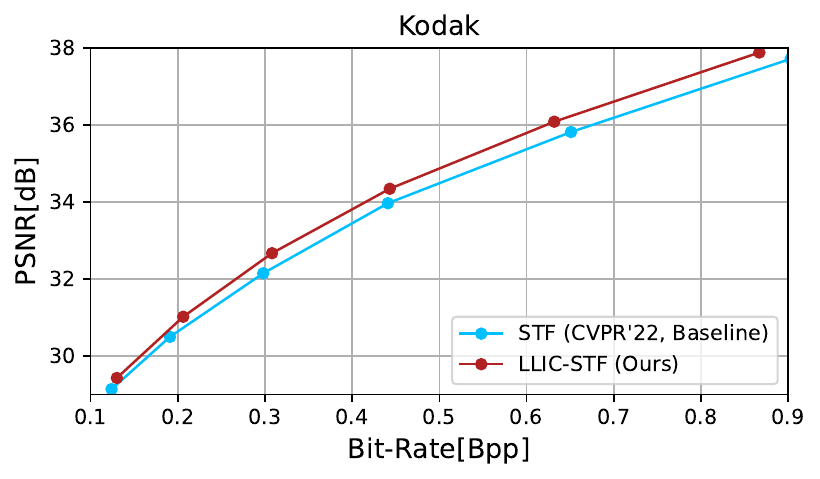}}
    \subfloat{
        \includegraphics[scale=0.43]{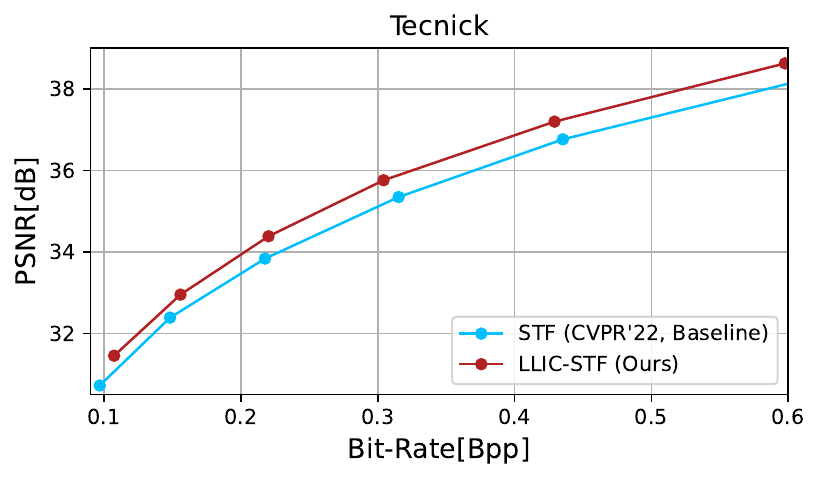}}
    \subfloat{
        \includegraphics[scale=0.43]{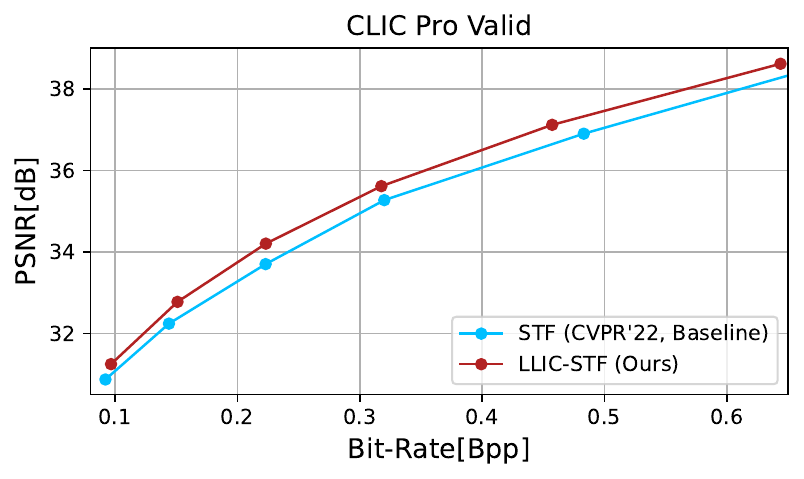}}\\
	\subfloat{
		\includegraphics[scale=0.43]{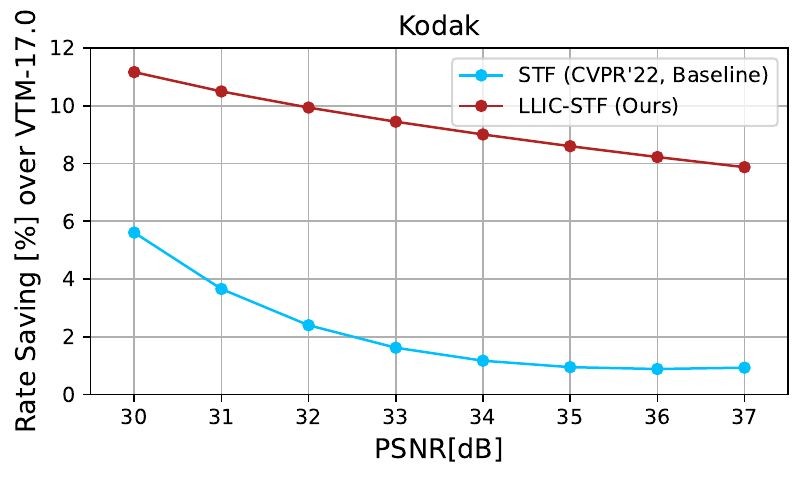}}
    \subfloat{
        \includegraphics[scale=0.43]{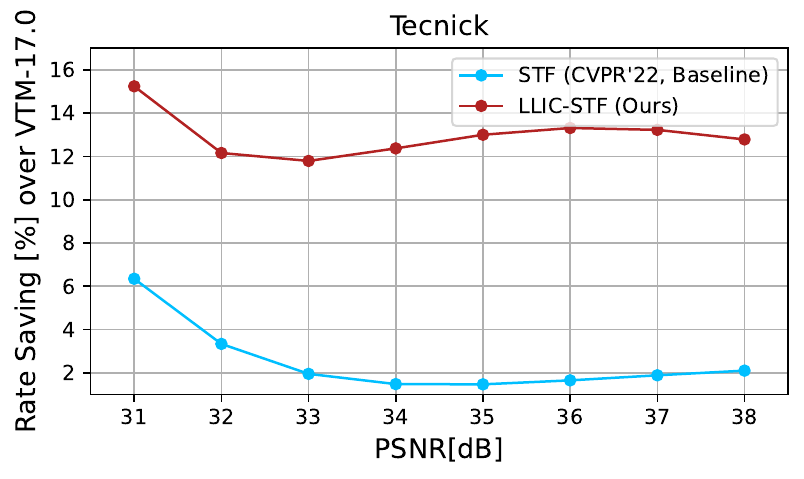}}
    \subfloat{
        \includegraphics[scale=0.43]{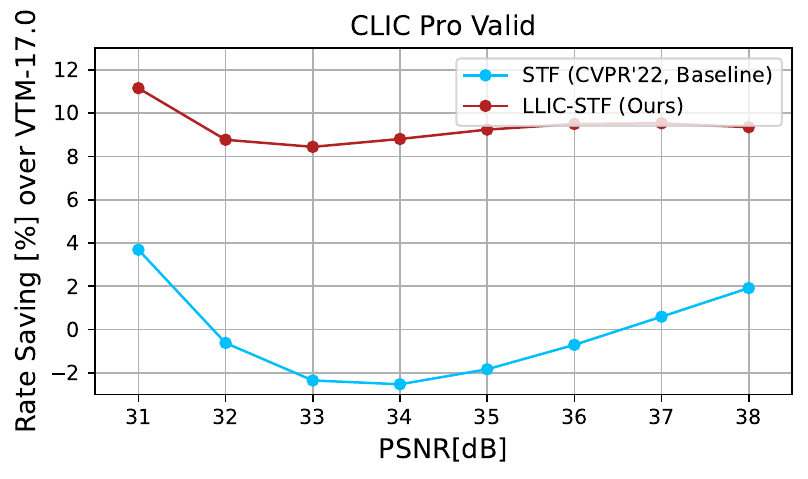}}
	\caption{PSNR-Bit-Rate curves and Rate saving-PSNR curves of our proposed LLIC-STF and its baseline STF~\cite{zou2022the}.
    The relative rate-saving curves are generated by first interpolating the discrete RD points with a cubic spline and then comparing the bitrates of different models at a fixed PSNR.}
	\label{fig:rd_stf}
\end{figure*}
The complexity of an FFN is approximately $O(4c^2hw + 2chw)$, while the complexity of the proposed
gate block is $O(3c^2hw+chw)$. The gate block is more efficient.
\subsection{Proposed Channel Transform Block (CTB)}\label{sec:method:ctb}
{In the proposed Spatial Transform Block (STB), the kernel weights are connected in a depth-wise manner,
necessitating the enhancement of interactions among channels.
Compared to baseline models and existing learned image compression approaches,
STB employs significantly larger kernels, enabling a more expansive receptive field at the same depth.
This increased receptive field capacity provides an opportunity to further strengthen the interactions between channels.}
Our proposed CTB is illustrated in Figure~\ref{fig:basic}.
The architecture of CTB is similar to that of STB. 
The SCST in STB is replaced with Self-Conditioned Channel Transform (SCCT)
to build the CTB. DepthRB for non-linear embedding and gate block are 
also employed. The overall process is formulated as
\begin{equation}
    \begin{aligned}
        \boldsymbol{\mathcal{F}}_{ebd}^{ctb} &= \textrm{DepthRB}(\textrm{Norm}(\boldsymbol{\mathcal{F}}_{in}^{ctb})),\\
        {\boldsymbol{\mathcal{F}}_{scst}^{ctb}} &= \textrm{SCCT}(\boldsymbol{\mathcal{F}}_{ebd}^{ctb}){,}\\
        {\boldsymbol{\mathcal{F}}_{skip}^{ctb}} &= \textrm{Conv}{1\times1}\left({\boldsymbol{\mathcal{F}}_{scst}^{ctb}}\right) +  \boldsymbol{\mathcal{F}}_{in}^{ctb},\\
        \boldsymbol{\mathcal{F}}_{out}^{ctb} &= \textrm{Gate}(\textrm{Norm}({\boldsymbol{\mathcal{F}}^{ctb}_{skip}})) + {\boldsymbol{\mathcal{F}}_{skip}^{ctb}},    \end{aligned}
\end{equation}
where $\boldsymbol{\mathcal{F}}_{in}^{ctb}$ is the input feature and $\boldsymbol{\mathcal{F}}_{out}^{ctb}$ is the output feature.
\subsubsection{Self-Conditioned Channel Transform (SCCT)}\label{sec:method:crb:ca}
In STB, large receptive field kernels with self-conditioned adaptability are employed to reduce spatial redundancy.
Due to the limited interactions among channels, Self-Conditioned Channel Transform (SCCT) is introduced
to reduce channel-wise redundancy.
{It is important to recognize that different channels within a feature map carry varying levels of information. 
Some channels contain crucial information for reconstruction, while others may be less informative.
By allocating more bits to critical channels and fewer bits to non-critical channels, we can achieve more efficient utilization of the available bit-rate.}
To address this issue and inspired by Channel Attention~\cite{hu2018squeeze}, we propose {generating} adaptive channel
factors to modify channel weights.
The proposed SCCT contains a condition branch and a main branch.
In the condition branch, the progressive down-sampling strategy is also adopted, which reduces information loss.
The input feature $\boldsymbol{\mathcal{F}}_{in}^{scct}$ is employed as a condition.
$\boldsymbol{\mathcal{F}}^{scct}_{in}$ is average pooled to reset the resolution
and {obtain} $\boldsymbol{\mathcal{F}}_{pool}^{scct}\in \mathbb{R}^{c\times 3\times 3}$.
The self-conditioned channel scaling factor {$\boldsymbol{\mathcal{W}}^{scct}\in \mathbb{R}^{c\times 1\times 1}$}
is computed via the convolution between $\boldsymbol{\mathcal{F}}_{pool}^{scct}$ and the weights of the condition branch.
The process is formulated as follows:
\begin{equation}
    \begin{aligned}
        \boldsymbol{\mathcal{F}}_{pool}^{scct} &= \textrm{AvgPool}(\boldsymbol{\mathcal{F}}_{in}^{scct}),\\
        {\boldsymbol{\mathcal{W}}^{scct}} &= \textrm{Conv}1\times1(\textrm{Conv}3\times3(\boldsymbol{\mathcal{F}}_{pool}^{scct})),\\
        \boldsymbol{\mathcal{F}}^{scct}_{out} &= {\boldsymbol{\mathcal{W}}^{scct}}\odot  \textrm{Conv}1\times1(\boldsymbol{\mathcal{F}}^{scct}_{in}),\\
    \end{aligned}
\end{equation}
where $\boldsymbol{\mathcal{F}}^{scct}_{out}$ is the output feature.
{Because} the channel factor is dependent {on} the condition, our proposed channel transform is adaptive.
\subsection{Improved Training Techniques}
In previous works, the kernel sizes are $3\times 3$ and $5\times 5$,
which is not sufficient for redundancy reduction. In our STB,
the kernel size is scaled up to $11\times 11$ and $9\times 9$, which
further enlarge the receptive field.
However, it is \textit{non-trivial} to fully exploit the potential of large kernels.
Vanilla training strategies (using $256\times 256$ patches) adopted
by previous works~\cite{cheng2020learned,zou2022the,he2022elic} cannot
fully utilize the large convolutional kernel.
When using $256\times 256$ patches, the resolutions of features during the analysis transform are
$\{128\times 128, 64\times 64, 32\times 32$, and $16\times 16\}$, which are too small,
especially $32\times 32$ and $16\times 16$ for the $11\times 11$ and $9\times 9$ kernels.
To address this issue, we propose training using $512\times 512$ patches.
{To mitigate the overhead, we adopt a two-stage training approach. First, we train the models on $256\times 256$
 patches to establish a solid foundation. Subsequently, we train the models on $512\times 512$ patches.   
 This allows them to refine their understanding of larger-scale spatial relationships and fully exploit the potential of the large kernels.}
This approach is quite simple yet effective.
\section{Experiments}\label{sec:exp}
\subsection{Implementation Details}\label{sec:exp:imple}
\subsubsection{Training dataset Preparation}
The proposed LLIC-STF, LLIC-ELIC, and LLIC-TCM are trained on $98939$ images
from COCO2017~\cite{lin2014microsoft}, ImageNet~\cite{deng2009imagenet}, DIV2K~\cite{Agustsson2017NTIRE2C}, 
and Flicker2K~\cite{lim2017enhanced}.
The initial resolutions of these training images are {greater} than $512\times 512$.
Following Ball{\'e} \textit{et al}~\cite{balle2018variational}, in order to reduce compression artifacts that may be present in JPEG format images, JPEG images are further down-sampled 
with a randomized factor using {the} PIL library.
This down-sampling process ensures that the minimum height or width of the images falls within the range of 512 to 584 pixels.
\begin{figure*}[t]
	\centering
	\subfloat{
		\includegraphics[scale=0.43]{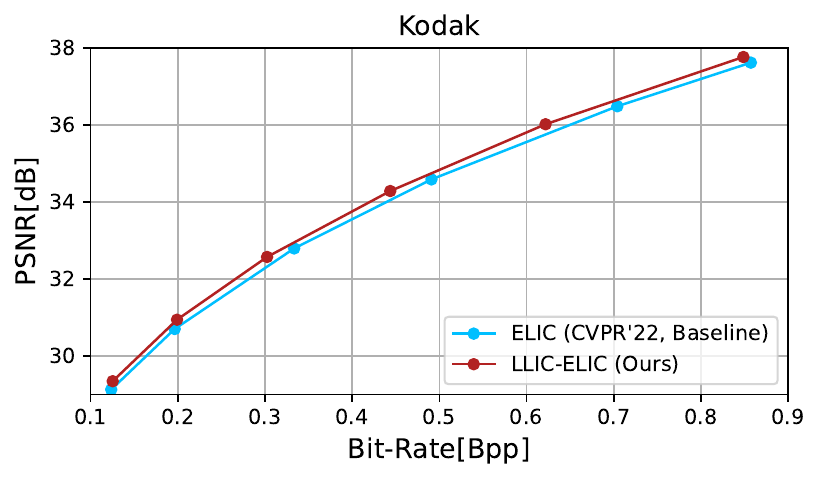}}
    \subfloat{
        \includegraphics[scale=0.43]{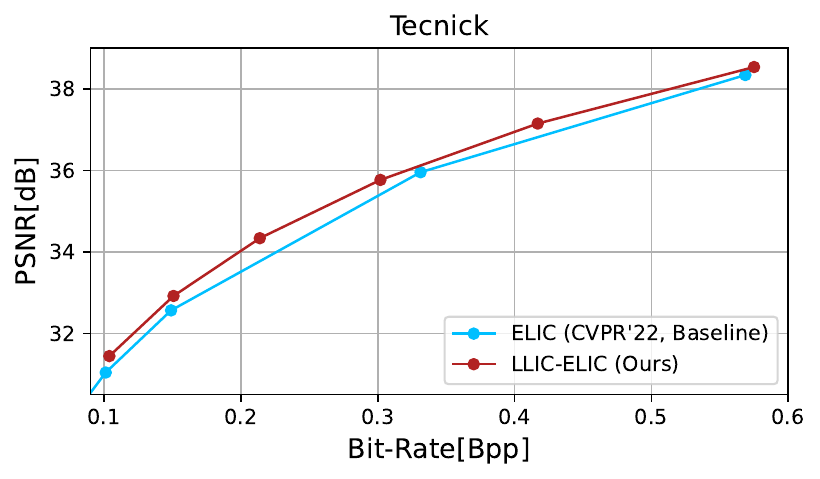}}
    \subfloat{
        \includegraphics[scale=0.43]{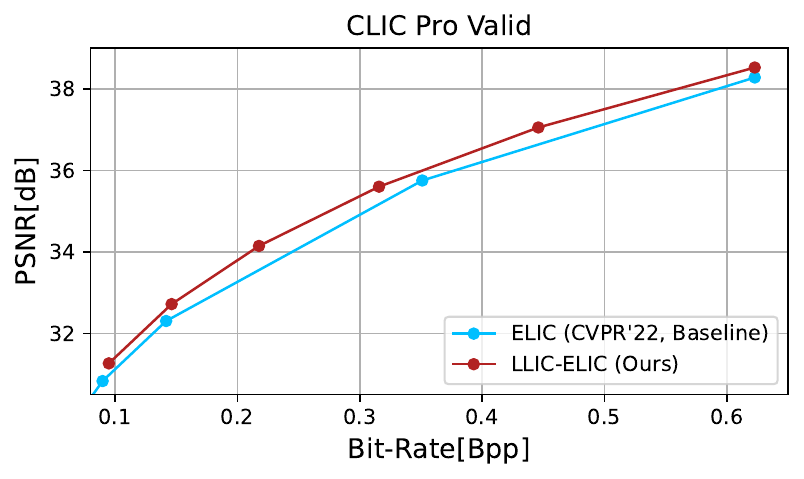}}\\
	\subfloat{
		\includegraphics[scale=0.43]{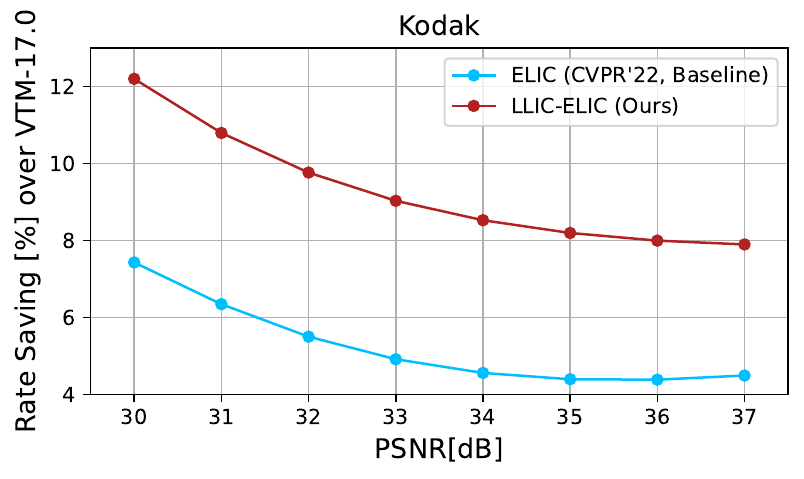}}
    \subfloat{
        \includegraphics[scale=0.43]{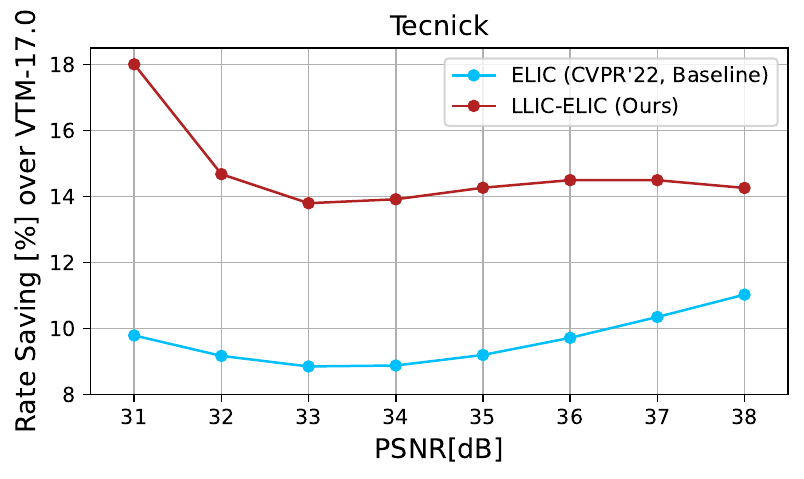}}	
    \subfloat{
        \includegraphics[scale=0.43]{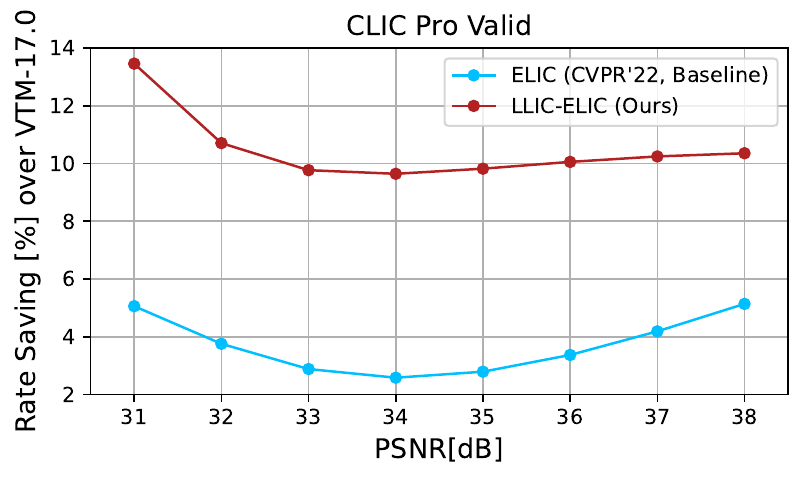}}
	\caption{PSNR-Bit-Rate curves and Rate saving-PSNR curves of our proposed LLIC-ELIC and its baseline ELIC~\cite{he2022elic}.}
	\label{fig:rd_elic}
\end{figure*}
\subsubsection{Training Strategy}\label{sec:exp:imple:train}
Our proposed LLIC-STF, LLIC-ELIC, and LLIC-TCM are built on PyTorch 2.1~\cite{paszke2019pytorch}
and CompressAI 1.2.0b3~\cite{begaint2020compressai}. 
Equation~\ref{eq:loss} is employed as the loss function.
Following existing methods~\cite{balle2018variational, minnen2018joint,minnen2020channel,cheng2020learned,he2022elic,jiang2022mlic},
the MSE and Multi-Scale Structural Similarity (MS-SSIM) are employed as 
distortion metrics during training for rate-distortion optimization.
Following CompressAI~\cite{begaint2020compressai}, $\lambda$ in Equation~\ref{eq:loss} is set to
$\{18, 35, 67, 130, 250, 483\}\times 10^{-4}$ for MSE and $\{2.4, 4.58, 8.73, 16.64, 31.73, 60.5\}$ for MS-SSIM.
The training process is conducted on $2$ Intel(R) Xeon(R) Platinum 8260 CPUs and $8$ Tesla V100-32G GPUs under the PyTorch distributed data parallel setting.
During training, the batch size is set to $16$.
Each model is trained for $2$ M steps.
The initial learning rate is $10^{-4}$. 
The learning rate {decreases} to $3\times 10^{-5}$ at $1.7$ M steps, {decreases} to $10^{-5}$
at $1.8$ M steps, {decreases} to $3\times 10^{-6}$ at $1.9$ M steps, and finally {decreases}
to $1.95$ M steps.
The training images are randomly cropped to $256\times 256$ patches during the first 1.2M steps and randomly cropped 
to $512\times 512$ patches during the {remaining} steps. Large patches are employed to
improve the performance of large receptive field transform coding.
\subsection{Test settings}\label{sec:exp:impel:test}
\subsubsection{Performance Test Settings}
For models optimized for MSE, Peak Signal-to-Noise Ratio (PSNR) {serves as the primary metric for quantifying distortion}.
The bits per pixel (Bpp) are utilized to measure bit-rate.
The rate-distortion performances are measure on $5$ widely used datasets, including
\begin{itemize}
    \item Kodak~\cite{kodak}, which contains $24$ uncompressed images, is widely used in
    learned image compression community~\cite{balle2016end,balle2018variational,ma2019iwave,akbari2021learned,minnen2018joint,cheng2020energy,yin2020co,mei2022scalable,minnen2020channel,cheng2020learned,qian2020learning,xie2021enhanced,zou2022the,zhu2022transformerbased,he2022elic,jiang2022mlic}.
    The resolution of the images in Kodak is $768\times 512$.
    \item Tecnick~\cite{tecnick2014TESTIMAGES}, which contains $100$ uncompressed images, is utilized for performance evaluation
    in many previous works~\cite{minnen2018joint,minnen2020channel,xie2021enhanced,liu2023learned,pan2022content}. The resolution of images
    in Tecnick is $1200\times 1200$.
    \item CLIC Pro Valid~\cite{CLIC2020} and CLIC 2021 Test~\cite{CLIC2021dataset} are the validation dataset of the 3rd Challenge on Learned Image Compression and
    the test dataset of the 4th Challenge on Learned Image Compression, respectively. CLIC Pro Valid contains $41$ high-resolution images, and 
    CLIC 2021 Test contains $60$ high-resolution images. Due to the impact of Challenge on Learned Image Compression, CLIC datasets are
    widely employed for rate-distortion evaluation~\cite{cheng2020learned,he2022elic,liu2023learned,zou2022the,zhu2022unified,feng2023nvtc}.
    The average resolution of images in CLIC datasets is approximately $2048\times 1370$.
    \item JPEG AI Test~\cite{jpegai} is the test dataset of the MMSP 2020 Learning-based Image Coding Challenge
    and contains $16$ images. The largest resolution
    of the images in JPEGAI Test is $3680\times 2456$.
\end{itemize}
These test datasets with different resolution{s} (from $768\times 512$ to $3680\times 2456$)
offer a comprehensive evaluation of learned image compression models.\par
The Bjøntegaard delta rate ({BD-Rate})~\cite{bjontegaard2001bdbr} is utilized to rank the performance of the learned image compression models.
\subsubsection{Complexity Test Settings}\label{sec:test:complex}
To comprehensively evaluate the complexities of various learned image compression models, $16$ images with resolutions {greater} than
$3584\times 3584$ from the LIU4K test split~\cite{liu2020comprehensive} are selected as the complexity test images.
These $16$ images are further center cropped to $\{512\times 512, 768\times 768, 1024\times 1024, 1536\times 1536, 2048\times 2048,
2560\times 2560, 3072\times 3072, 3584\times 3584\}$ patches to
cover the various resolutions of images that can be encountered in reality.
The model complexities are evaluated from four perspectives, including the peak GPU memory during encoding and decoding,
model Forward inference MACs, encoding time, and decoding time. The encoding and decoding times include the entropy coding and decoding times
to better match practical applications and realistic scenarios.
\subsection{Rate-Distortion Performance}\label{sec:exp:perf}
\subsubsection{Quantitative Results}
\begin{figure*}[t]
	\centering
	\subfloat{
		\includegraphics[scale=0.43]{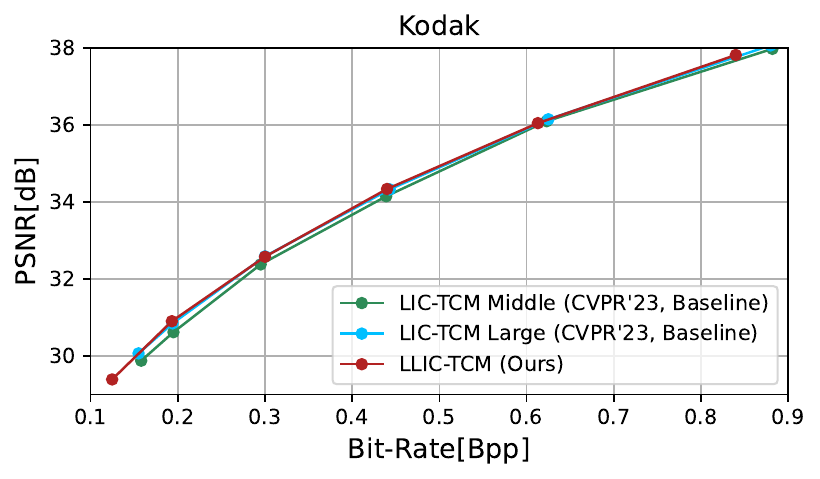}}
    \subfloat{
        \includegraphics[scale=0.43]{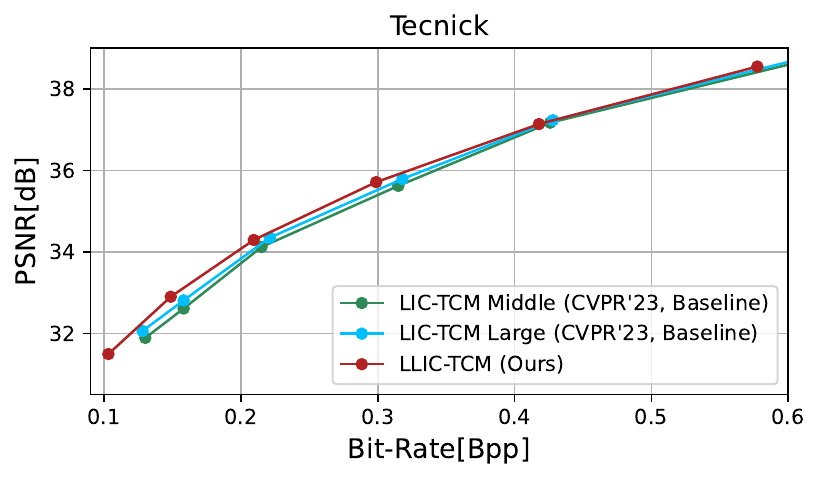}}
    \subfloat{
        \includegraphics[scale=0.43]{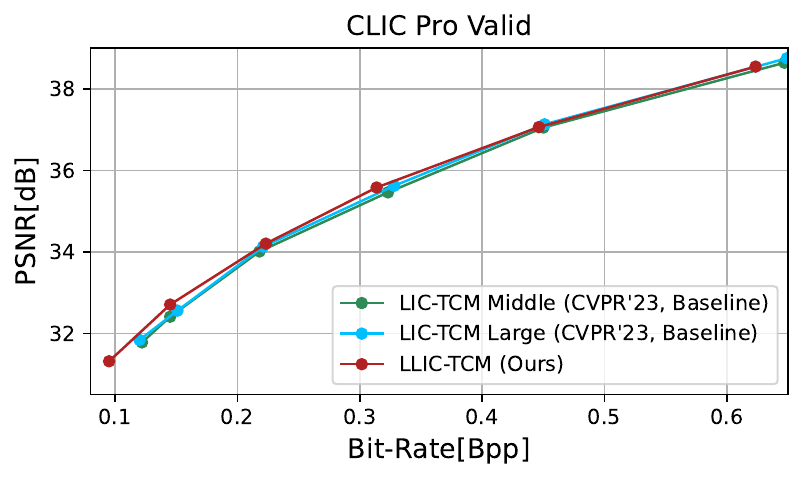}}\\
	\subfloat{
		\includegraphics[scale=0.43]{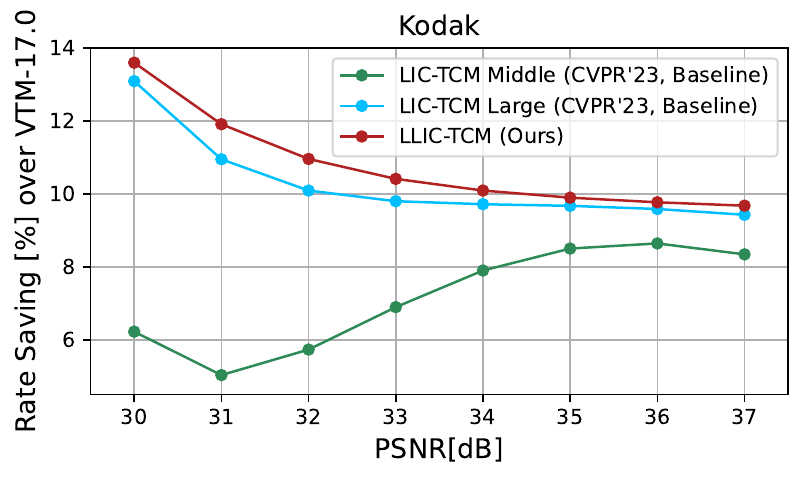}}
    \subfloat{
        \includegraphics[scale=0.43]{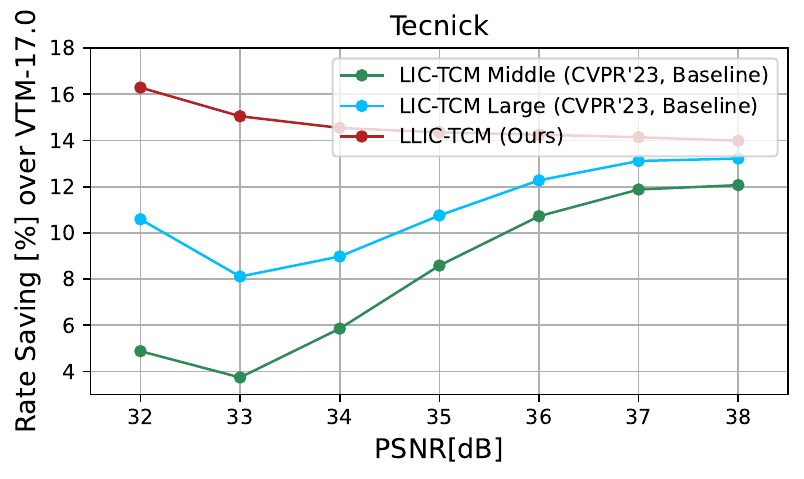}}
    \subfloat{
        \includegraphics[scale=0.43]{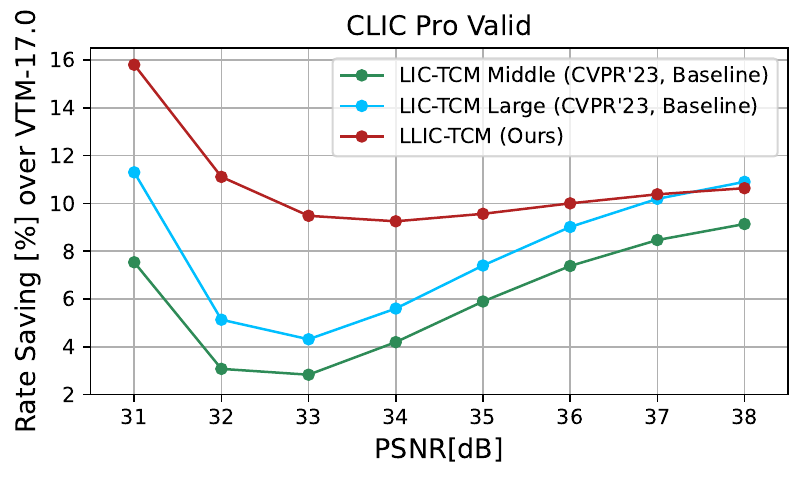}}
	\caption{PSNR-Bit-Rate curves and Rate saving-PSNR curves of our proposed LLIC-TCM and its baseline LIC-TCM~\cite{liu2023learned}.
    We highlight the performance improvements over LIC-TCM Middle because the MACs of our LLIC-TCM is $321.93$ G and the MACs of
    LIC-TCM Middle is $415.2$ G on Kodak. The MACs of LIC-TCM Large is $717.08$ G on Kodak. Our LLIC-TCM also outperforms 
    LIC-TCM Large with only half the MACs of LIC-TCM Large.}
	\label{fig:rd_tcm}
\end{figure*}

We compare our proposed LLIC-STF, LLIC-ELIC, and LLIC-TCM with their
baseline models STF~\cite{zou2022the}, ELIC~\cite{he2022elic}, and LIC-TCM~\cite{liu2023learned}, 
recent learned image compression models~\cite{cheng2020learned,minnen2020channel,qian2020learning,xie2021enhanced,qian2022entroformer,zhu2022transformerbased,wang2022neural,zhu2022unified,koyuncu2022contextformer,pan2022content,feng2023nvtc},
and the non-learned image codec VTM-17.0 Intra~\cite{bross2021vvc}.
The rate-distortion curves and rate-savings-distortion curves are illustrated in Fig.~\ref{fig:rd_stf},
Fig.~\ref{fig:rd_elic}, and Fig.~\ref{fig:rd_tcm}. 
The {BD-Rate} reductions are presented in Table~\ref{tab:rd}.
VTM-17.0 under the \textit{encoder\_intra\_vtm.cfg} in the YUV444 color space is employed as an anchor.
\par
Compared with {the} baseline model STF~\cite{zou2022the}, 
{our LLIC-STF achieves superior performance across all bit-rate tiers.
Specifically, LLIC-STF achieves average enhancements of $0.33, 0.47, 0.40, 0.50,$ and $0.62$ dB in PSNR over STF
across Kodak~\cite{kodak}, Tecnick~\cite{tecnick2014TESTIMAGES}, CLIC Pro Valid~\cite{CLIC2020}, CLIC 2021 Test, and JPEGAI Test~\cite{jpegai}, respectively.}
Our LLIC-STF {decreases the bitrate} by $7.2\%, 10.88\%, 9.81\%, 11.01\%, 12.34\%$ on these datasets when the anchor is STF.\par
Compared with baseline model ELIC~\cite{he2022elic},
{LLIC-ELIC shows markedly enhanced efficacy across all evaluated bit-rates.}
Our proposed LLIC-ELIC {yields} average improvements of $0.19, 0.25, 0.29, 0.21, 0.39$ dB
PSNR compared to ELIC on Kodak~\cite{kodak}, Tecnick~\cite{tecnick2014TESTIMAGES}, CLIC Pro Valid~\cite{CLIC2020}, CLIC 2021 Test, and JPEGAI Test~\cite{jpegai}, respectively.
Our LLIC-ELIC {decreases the bitrate} by $4.25\%, 5.66\%, 7.11\%, 4.80\%$, {and} $8.54\%$ on these datasets when the anchor is ELIC.\par
When comparing LLIC-TCM with LIC-TCM models, 
{it is pertinent to highlight the significant performance uplift over the LIC-TCM Middle,
noting that} the MACs of our LLIC-TCM is $321.93$ G and the MACs of the 
LIC-TCM Middle is $415.2$ G. The MACs of LIC-TCM Large is $717.08$ G.
Our LLIC-TCM performs much better than LIC-TCM Middle and slightly better than LLIC-TCM Large on Kodak. Our proposed LLIC-TCM {decreases the bitrate} by
$3\%$ more than LIC-TCM Middle and $1\%$ more than LIC-TCM Large.
Our LLIC-TCM performs much better than LIC-TCM models on \textit{high-resolution} images.
Our LLIC-TCM {decreases the bitrate} by $6\%, 4\%$ {more} than LIC-TCM Middle and $3.5\%$, and $2.4\%$
more than LIC-TCM Large on Tecnick and CLIC Pro Val, respectively. Our LLIC-TCM outperforms 
LIC-TCM Large with only half the MACs of LIC-TCM Large.
\begin{table*}[h]   
    \centering
    \small
    \setlength{\tabcolsep}{1mm}{
    \caption{{BD-Rate $(\%)$ comparison for PSNR (dB) and MS-SSIM. The anchor is VTM-17.0 Intra.}}
    \label{tab:rd} 
    \renewcommand\arraystretch{1.3}
    \begin{tabular}{@{}cccccccccccccc@{}}
    \toprule
    \multicolumn{1}{c}{\multirow{3}{*}{Methods}} & \multicolumn{1}{c|}{\multirow{3}{*}{{Venue}}}                            & \multicolumn{6}{c}{{BD-Rate (\%) w.r.t. {VTM-17.0 Intra}}}  \\
    \multicolumn{1}{c}{}      & \multicolumn{1}{c|}{}                                                & \multicolumn{2}{c}{Kodak~\cite{kodak}}    & \multicolumn{1}{c}{Tecnick~\cite{tecnick2014TESTIMAGES}}  & \multicolumn{1}{c}{CLIC Pro Valid~\cite{CLIC2020}} & \multicolumn{1}{c}{CLIC 2021 Test~\cite{CLIC2021dataset}} & \multicolumn{1}{c}{JPEGAI Test~\cite{jpegai}}  \\
    \multicolumn{1}{c}{}     & \multicolumn{1}{c|}{}                                                   & \multicolumn{1}{c}{PSNR} & \multicolumn{1}{c}{MS-SSIM}& \multicolumn{1}{c}{PSNR} & \multicolumn{1}{c}{PSNR} & \multicolumn{1}{c}{PSNR}  &\multicolumn{1}{c}{PSNR} \\ \midrule
    \multicolumn{1}{c}{Cheng'20~\cite{cheng2020learned}}     & \multicolumn{1}{c|}{CVPR'20}                                  & $+5.58$       & $-44.21$ & $+7.57$       & {$+11.71$} & {$+9.40$}       & {$+11.95$}       \\
    \rowcolor[HTML]{E6E6E6}\multicolumn{1}{c}{Minnen'20~\cite{minnen2020channel}}    & \multicolumn{1}{c|}{ICIP'20}                                       & $+3.23$       & $-$ & $-0.88$       & $-$ & $-$       & $-$       \\
    \multicolumn{1}{c}{Qian'21~\cite{qian2020learning}}      & \multicolumn{1}{c|}{ICLR'21}                                     & $+10.05$       & $-39.53$ & $+7.52$       & $+0.02$ & $-$       & $-$       \\
    \rowcolor[HTML]{E6E6E6}\multicolumn{1}{c}{Xie'21~\cite{xie2021enhanced}}        & \multicolumn{1}{c|}{ACMMM'21}                                   & $+1.55$       & $-43.39$ &  $-0.80$    & $+3.21$   & $+0.99$       & $+2.35$       \\
    \multicolumn{1}{c}{Entroformer~\cite{qian2022entroformer}}   & \multicolumn{1}{c|}{ICLR'22}                                       & $+4.73$       & $-42.64$ & $+2.31$       & $-1.04$ & $-$       & $-$       \\
    \rowcolor[HTML]{E6E6E6}\multicolumn{1}{c}{SwinT-Charm~\cite{zhu2022transformerbased}}    & \multicolumn{1}{c|}{ICLR'22}              & $-1.73$      & $-$   & $-$& $-$& $-$& $-$  \\
    \multicolumn{1}{c}{NeuralSyntax~\cite{wang2022neural}}   & \multicolumn{1}{c|}{CVPR'22}               & $+8.97$      & $-39.56$   & $-$& $+5.64$&$-$&  {$-$}  \\
    \rowcolor[HTML]{E6E6E6}\multicolumn{1}{c}{McQuic~\cite{zhu2022unified}}    & \multicolumn{1}{c|}{CVPR'22}            & $-1.57$      & $-47.94$   & $-$& $+6.82$&$-$&  {$-$}  \\
    \multicolumn{1}{c}{Contextformer~\cite{koyuncu2022contextformer}}  & \multicolumn{1}{c|}{ECCV'22}      & $-5.77$      & $-46.12$ & $-9.05$      & {$-$}  & $-$      & $-$     \\
    \rowcolor[HTML]{E6E6E6}\multicolumn{1}{c}{Pan'22~\cite{pan2022content}}  &  \multicolumn{1}{c|}{ECCV'22}      & $+7.56$      & $-36.2$ & $+3.97$      & $-$  & $-$      & $-$     \\
    \multicolumn{1}{c}{NVTC~\cite{feng2023nvtc}} & \multicolumn{1}{c|}{CVPR'23}       & $-1.04$      & $-$ & $-$      & $-$  & $-3.61$      & $-$     \\\midrule
    \textit{\textbf{STF as Baseline}} \\
    \multicolumn{1}{c}{STF~\cite{zou2022the}}   & \multicolumn{1}{c|}{CVPR'22}       & $-2.48$      & $-47.72$  & $-2.75$      & $+0.42$   & $-0.16$   & $+1.54$      \\          
    \rowcolor[HTML]{E6E6E6}\multicolumn{1}{c}{LLIC-STF}     & \multicolumn{1}{c|}{Ours} & {$\bm{-9.49}$}      & $\bm{-49.11}$ & $\bm{-13.06}$      & $\bm{-9.32}$  & $\bm{-11.44}$      & $\bm{-11.15}$     \\\midrule
    \textit{\textbf{ELIC as Baseline}} \\
    \multicolumn{1}{c}{ELIC~\cite{he2022elic}}         & \multicolumn{1}{c|}{CVPR'22}                     & $-5.95$      & $-44.60$   & $-9.14$      & $-3.45$ & $-7.52$      & $-3.21$   \\
    \rowcolor[HTML]{E6E6E6}\multicolumn{1}{c}{LLIC-ELIC  }     & \multicolumn{1}{c|}{Ours}   &{$\bm{-9.47}$}      & $\bm{-49.25}$ & $\bm{-14.68}$      & $\bm{-10.35}$  & $\bm{-12.32}$      & $\bm{-11.24}$     \\\midrule
    \textit{\textbf{LIC-TCM as Baseline}} \\
    \multicolumn{1}{c}{LIC-TCM Middle~\cite{liu2023learned}}  & \multicolumn{1}{c|}{CVPR'23}       & $-7.43$      & $-$ & $-8.99$      & $-6.35$  & $-$      & $-$     \\
    \multicolumn{1}{c}{LIC-TCM Large~\cite{liu2023learned}}   & \multicolumn{1}{c|}{CVPR'23}      & $-10.14$      & $-48.94$ & $-11.47$      & $-8.04$  & $-$      & $-$     \\
    \rowcolor[HTML]{E6E6E6}\multicolumn{1}{c}{LLIC-TCM }      & \multicolumn{1}{c|}{Ours}   & $\bm{-10.94}$      & $\bm{-49.73}$ & $\bm{-14.99}$      & $\bm{-10.41}$  & $\bm{-13.14}$      & $\bm{-12.30}$     \\\bottomrule
    \end{tabular}}
  \end{table*}
    \begin{figure*}[!t]
        \centering
        \includegraphics[width=\linewidth]
        {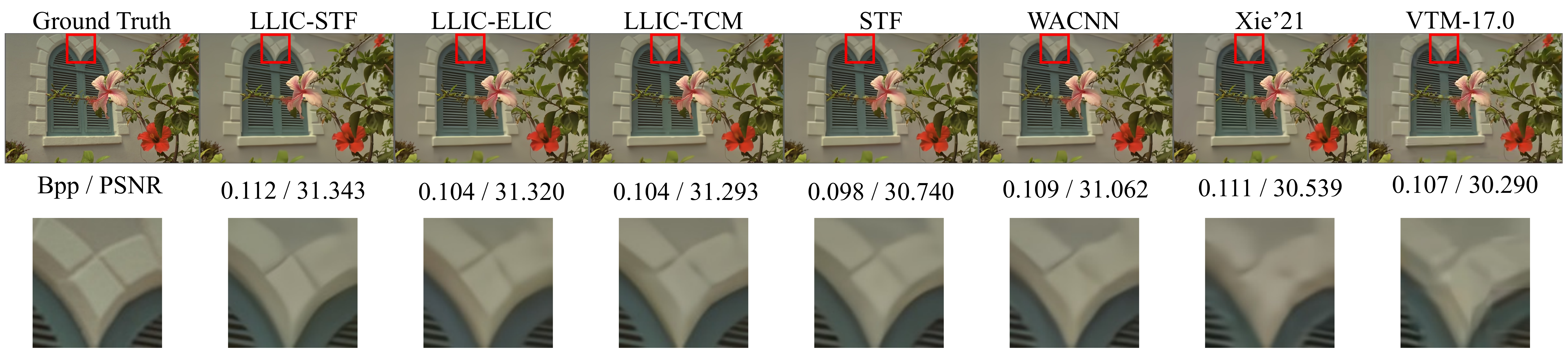}
        \caption{Visualization of the reconstructed Kodim07 from the Kodak dataset. The metrics are [bpp↓/PSNR↑].
        We compare our LLIC-STF, LLIC-ELIC, and LLIC-TCM with {STF~\cite{zou2022the}, WACNN~\cite{zou2022the},} Xie'21~\cite{xie2021enhanced}, and VTM-17.0 Intra~\cite{bross2021vvc}.}
        \label{fig:vis}
    \end{figure*}
    \begin{figure*}
        \centering
        \includegraphics[width=0.97\linewidth]
        {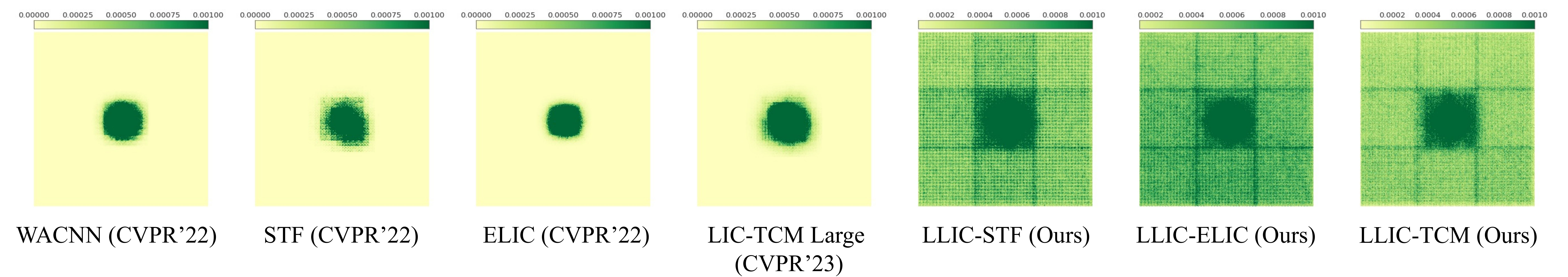}
        \caption{Effective Receptive Fields (ERF) of analysis transforms $g_a$ of
        WACNN, STF, ELIC, LIC-TCM Large and our proposed models on 24 Kodak images center-cropped to $512\times 512$.
        The ERF is visualized as the absolute gradients of the center pixel in the latent ($\mathrm{d}\boldsymbol{y}/\mathrm{d}\boldsymbol{x}$) with respect
        to the input image. {Darker green colors represent larger gradients. A more widely distributed green area indicates a larger ERF. {A} larger ERF indicates that
        more spatial contextual information is captured and utilized during the transform}.}
        \label{fig:erf}
    \end{figure*}
    \begin{figure}[h]
        \centering
        \subfloat[STF]{\includegraphics[width=.49\columnwidth]{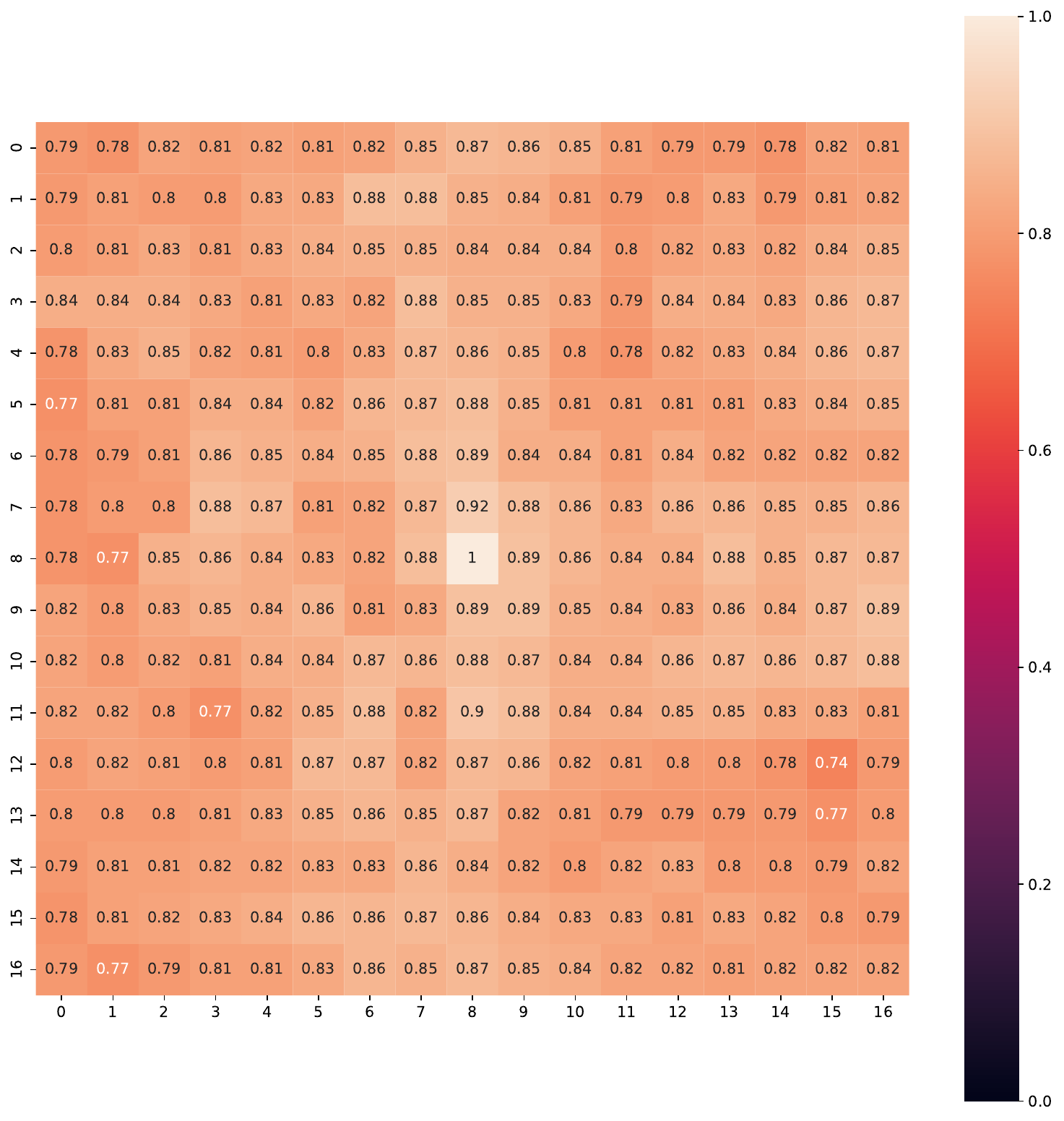}}
        \subfloat[ELIC]{\includegraphics[width=.49\columnwidth]{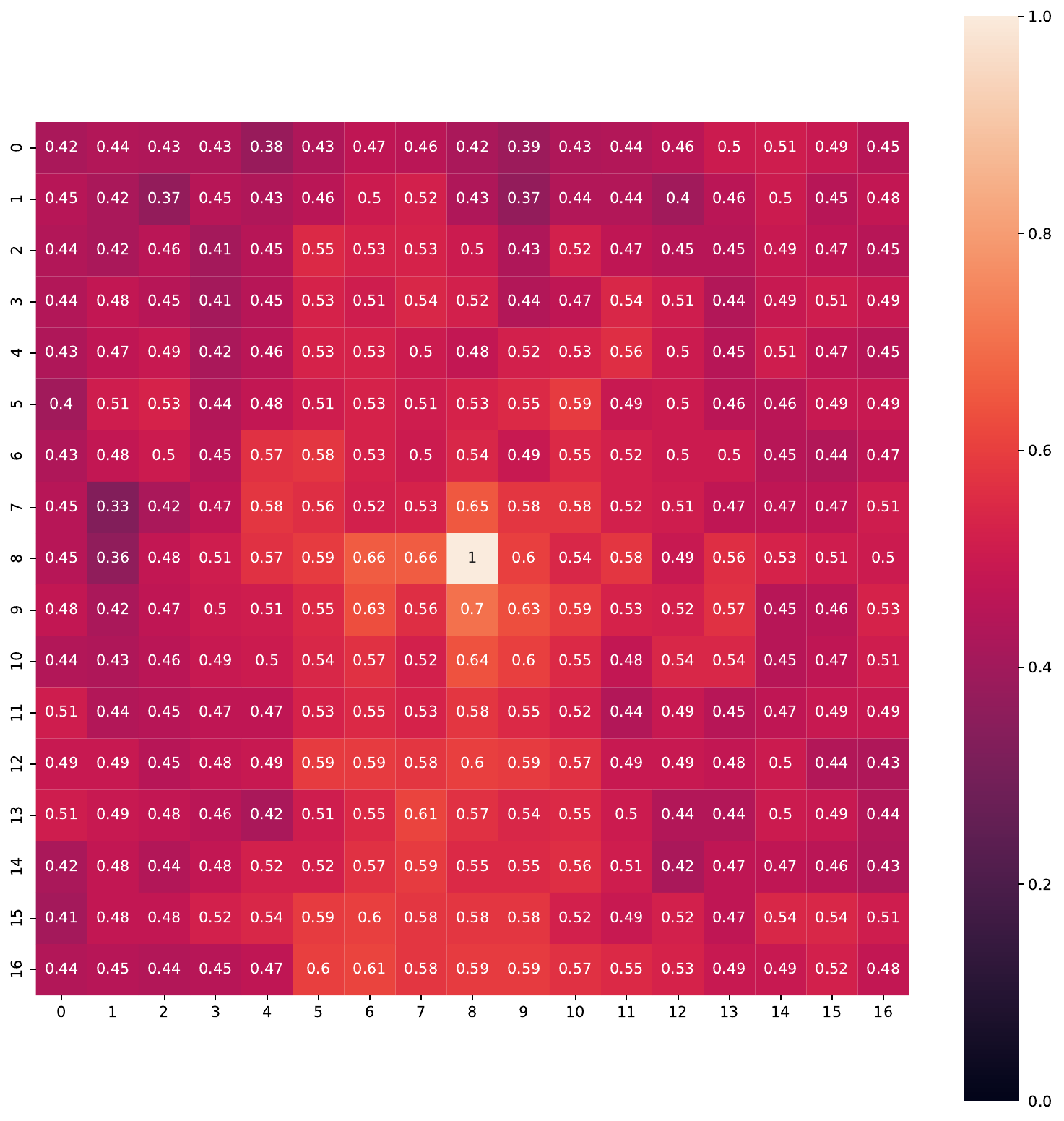}}\\
        \subfloat[LIC-TCM Large]{\includegraphics[width=.49\columnwidth]{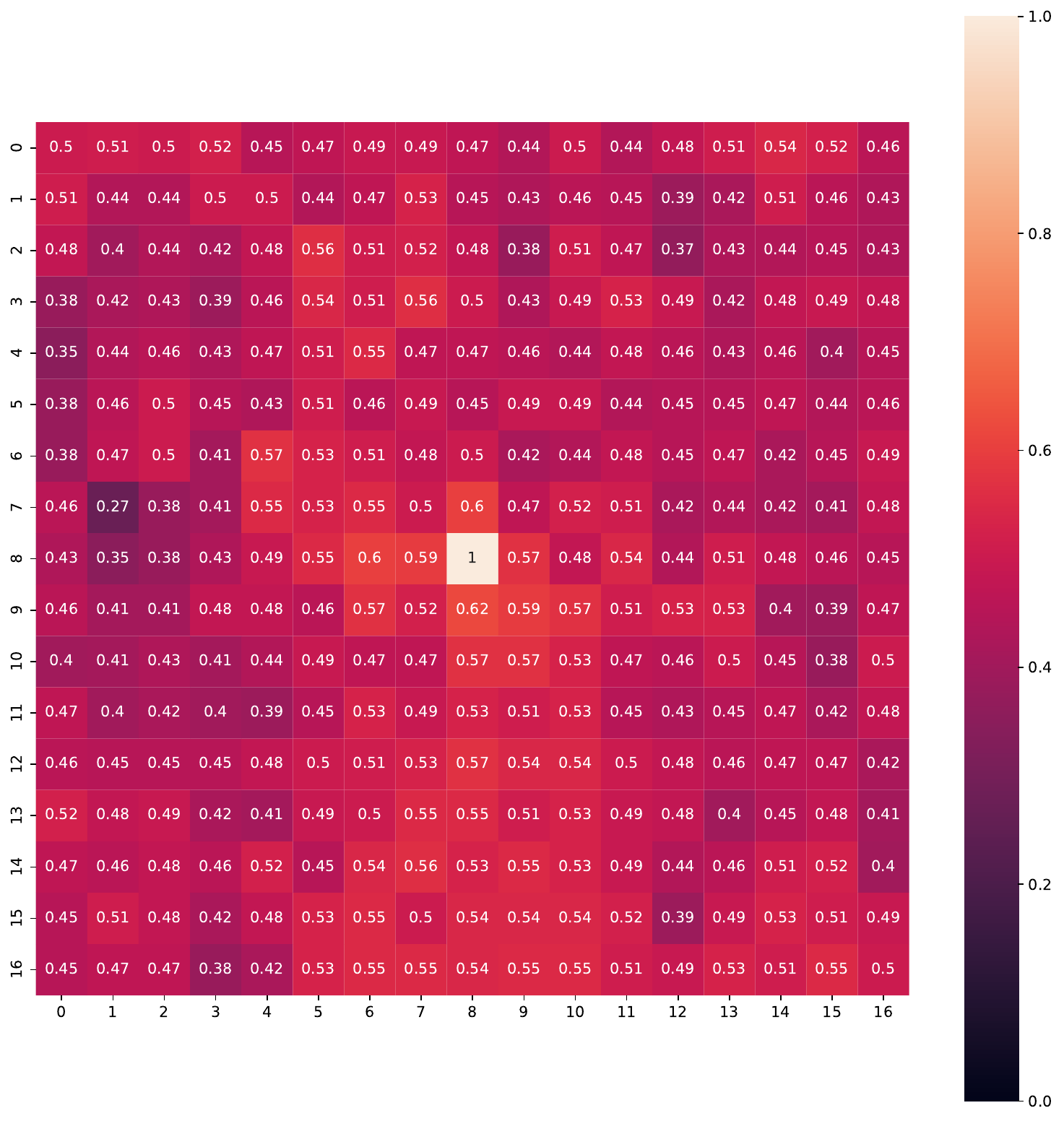}}
        \subfloat[LLIC-STF]{\includegraphics[width=.49\columnwidth]{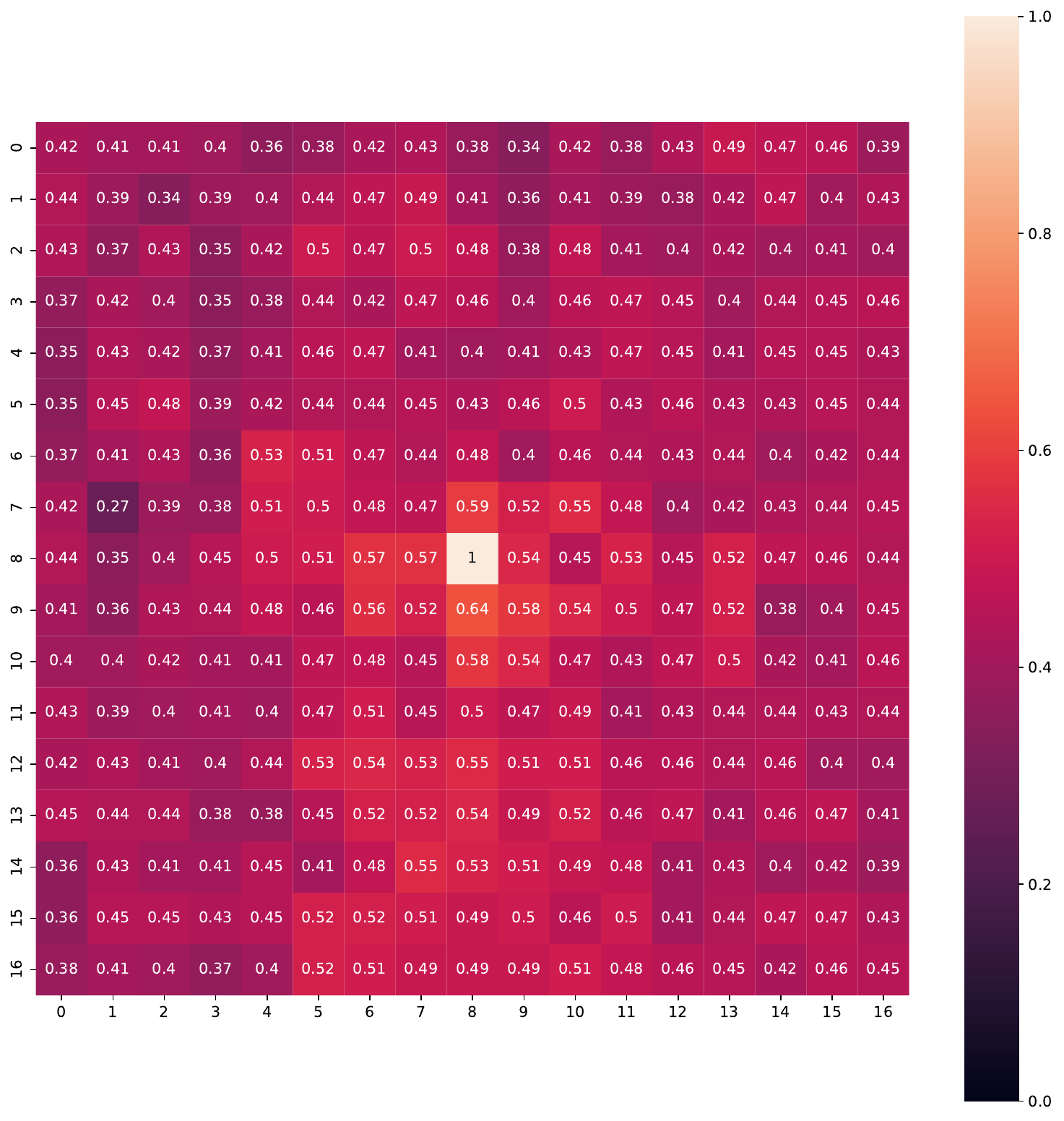}}\\
        \subfloat[LLIC-ELIC]{\includegraphics[width=.49\columnwidth]{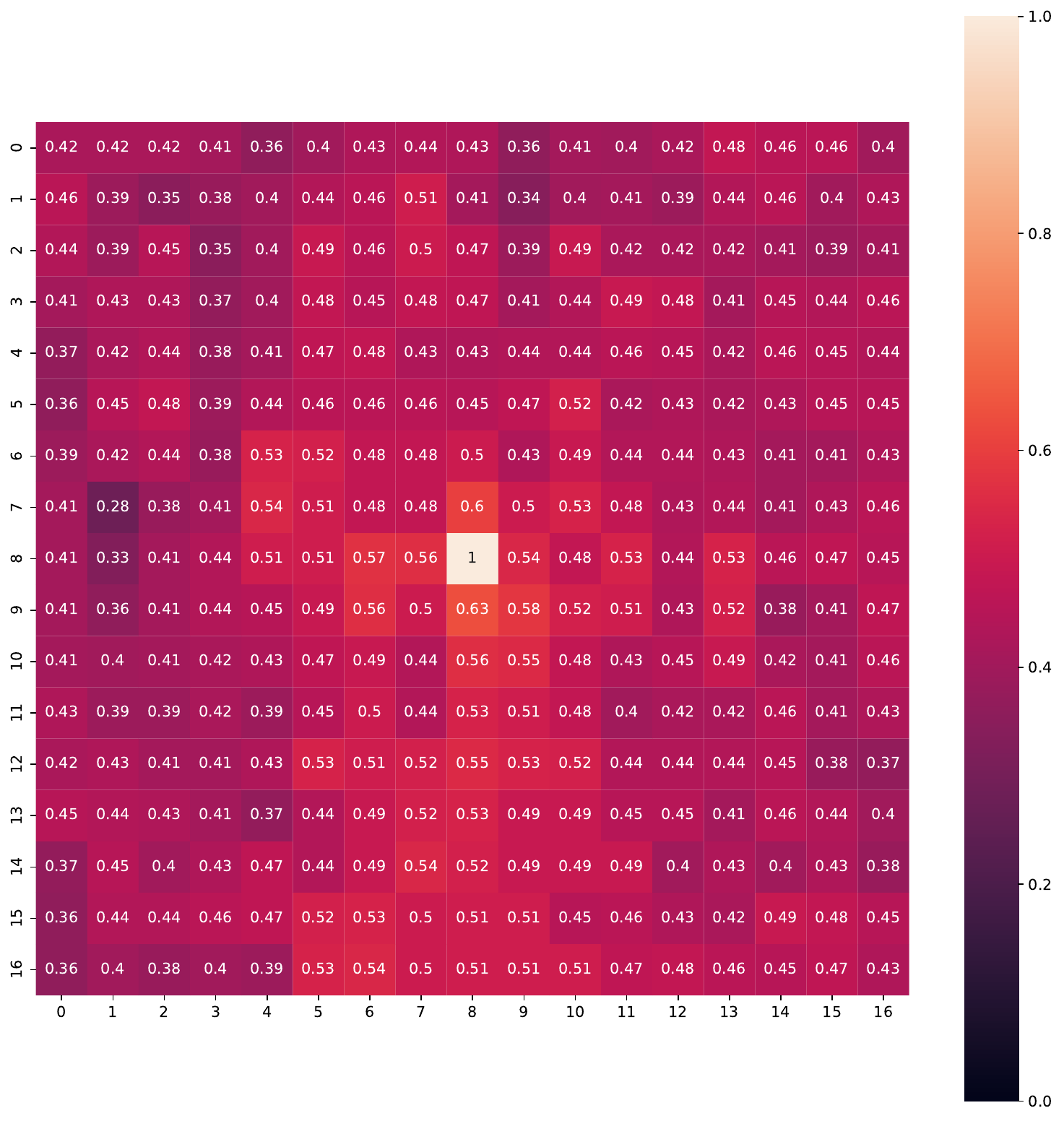}}
        \subfloat[LLIC-TCM]{\includegraphics[width=.49\columnwidth]{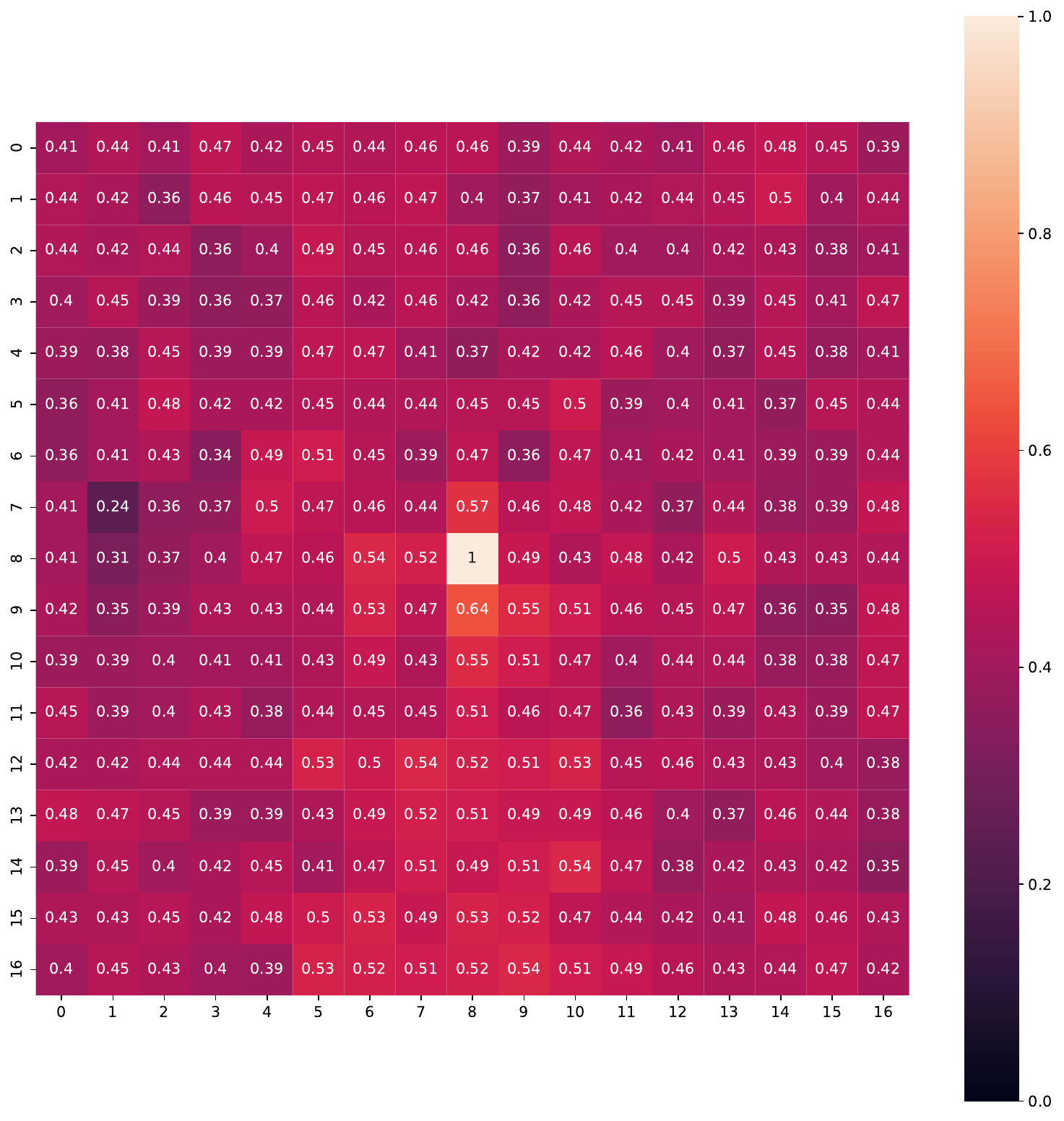}}\\
        \caption{{{Absolute value of the cosine similarity among latents}. Note that the forward MACs of our LLIC models are only half of the forward MACs of LIC-TCM Large. Please zoom in for {a} better view.}}
        \label{fig:cs}
      \end{figure}
    \begin{figure*}
        \centering
        \subfloat{
            \includegraphics[scale=0.41]{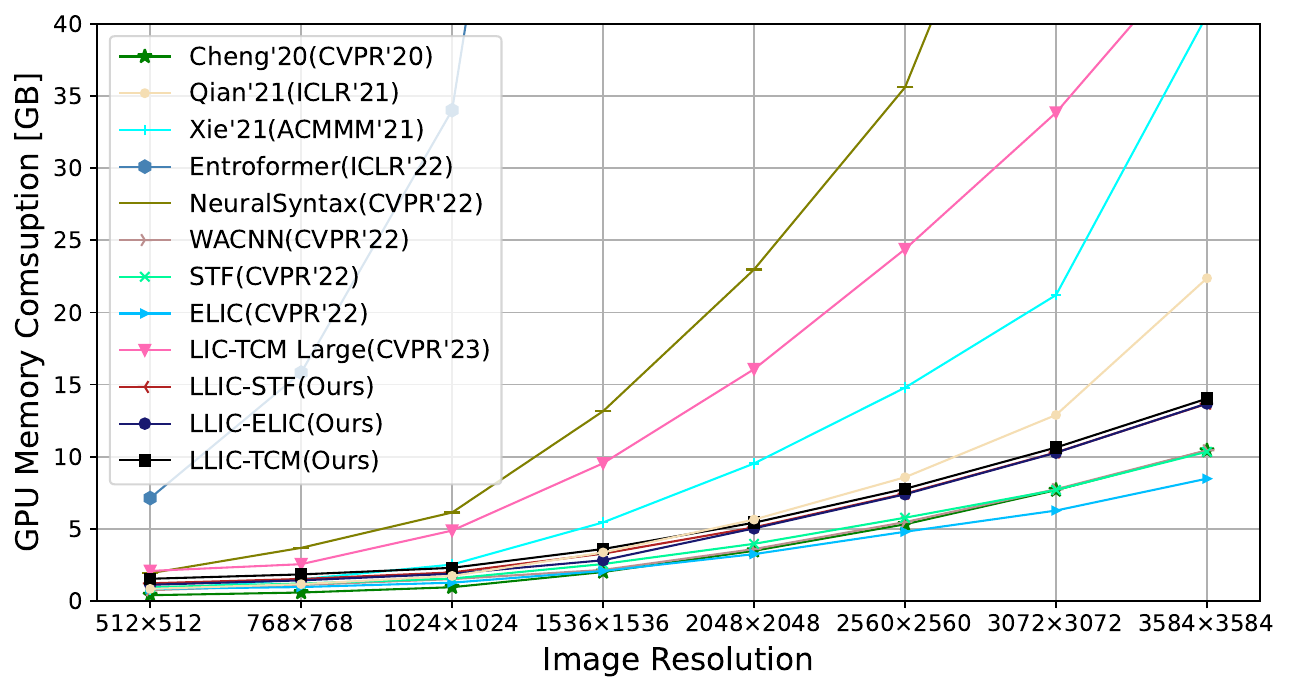}}
        \subfloat{
            \includegraphics[scale=0.41]{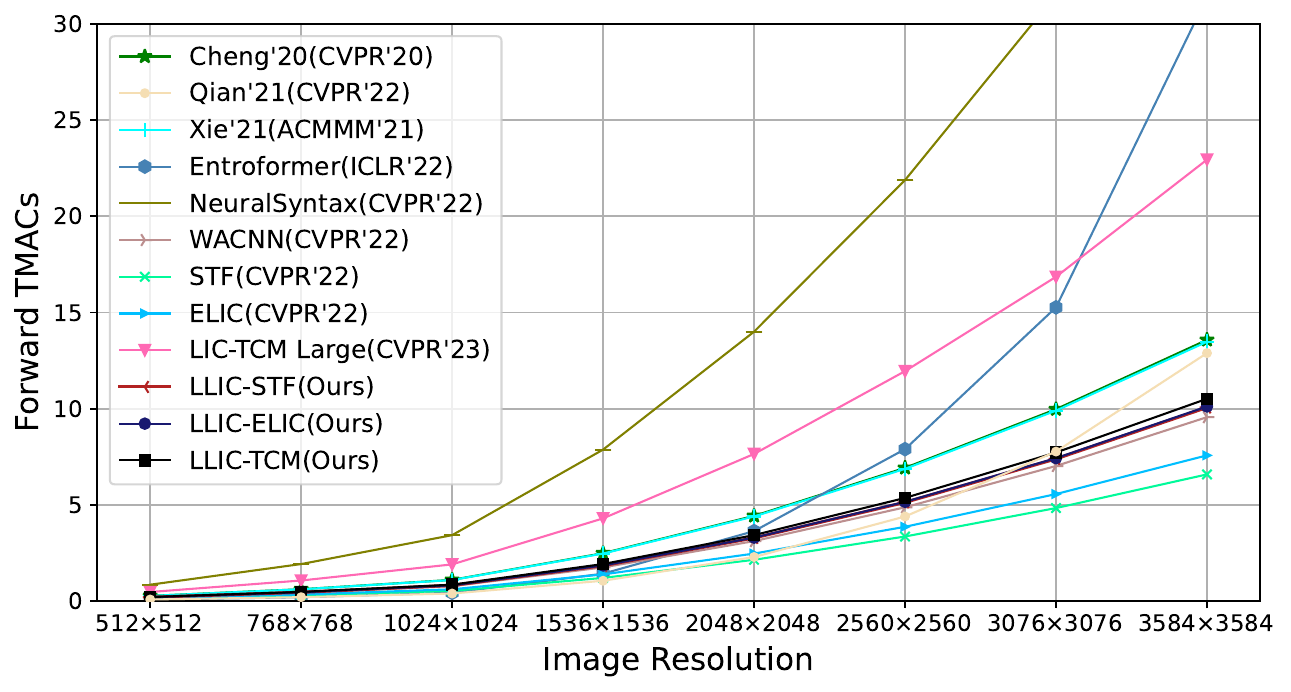}}\\
        \subfloat{
            \includegraphics[scale=0.41]{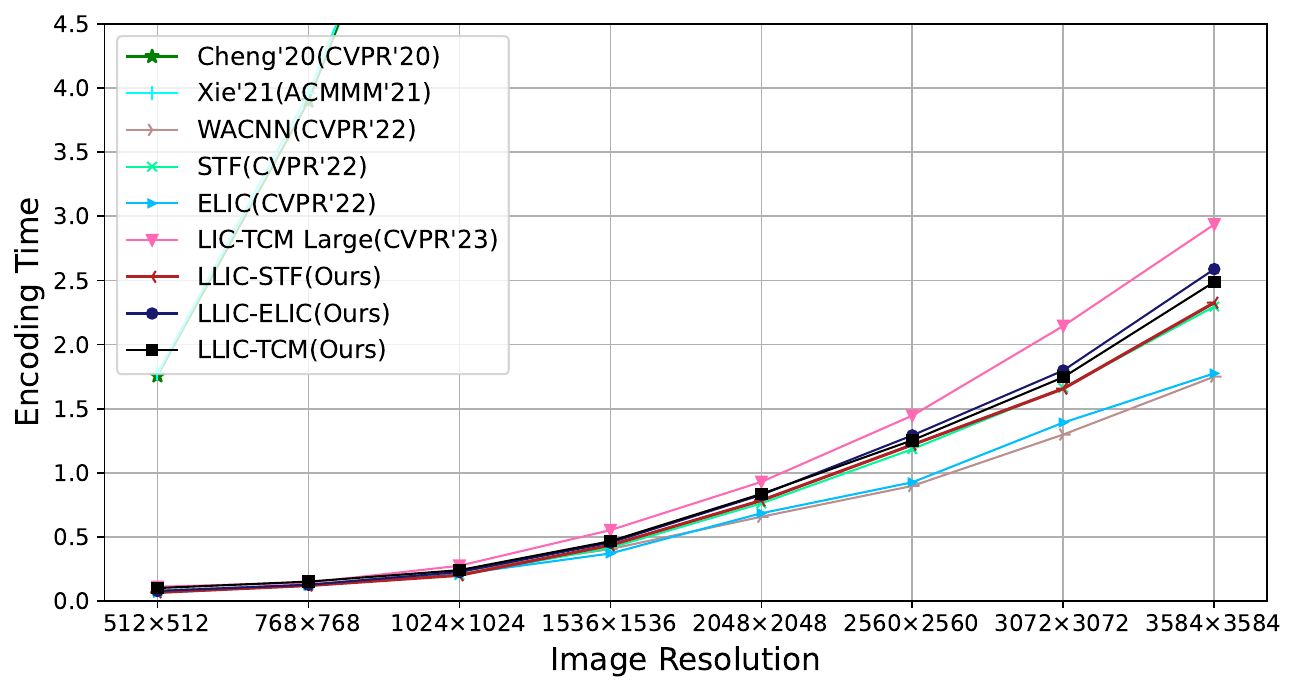}}
        \subfloat{
            \includegraphics[scale=0.41]{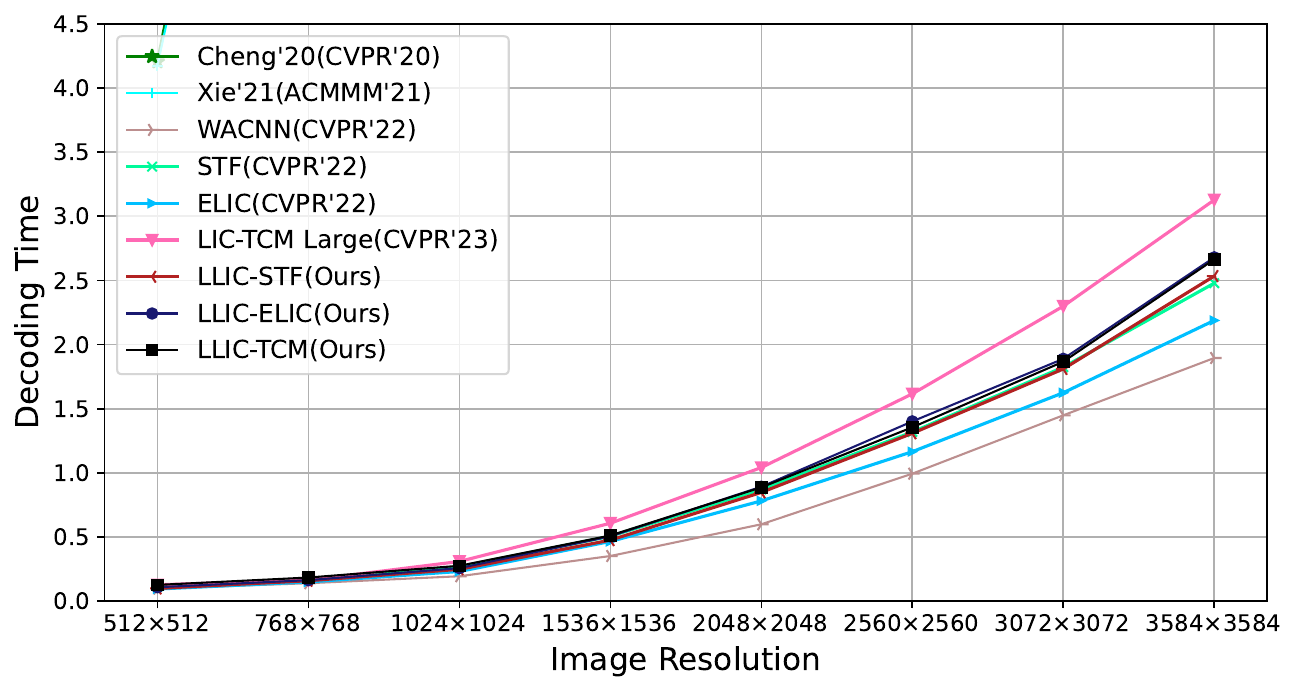}}
        \caption{GPU Memory Consumption-Image Resolution, {Forward MACs-Image Resolution}, Encoding Time-Image Resolution, Decoding Time-Image Resolution curves.}
        \label{fig:complex}
    \end{figure*}
        \begin{table}[t!]
            \centering
            \setlength{\tabcolsep}{7pt}
            \renewcommand{\arraystretch}{1.3}
            \caption{Ablation Studies on Kodak dataset.} 
            \label{tab:ablation}
            \setlength{\abovecaptionskip}{2pt}
            \begin{tabular}{cccc}
            \toprule
            \multicolumn{1}{c|}{Settings} & \multirow{1}{*}{BD-Rate (\%)}  & \multirow{1}{*}{Forward MACs (G)}\\
            \midrule
            \multirow{1}{*}{\textbf{LLIC-STF Basic Block}}\\
            \multicolumn{1}{c|}{STB + STB}                   & -6.05           &331.55                          \\
            \multicolumn{1}{c|}{CTB + CTB}  & -6.19  & 318.61\\
            \multicolumn{1}{c|}{only CTB}                   & -5.05           &214.90                          \\
            \multicolumn{1}{c|}{only STB}  & -4.19  & 221.29\\
            \midrule
            \multirow{1}{*}{\textbf{LLIC-STF w/o CTB}}                   \\
            \multicolumn{1}{c|}{static weight} & -2.58 & 221.26 \\
            \multicolumn{1}{c|}{w/o DepthRB}                    & -1.01         & 188.09            \\
            \multicolumn{1}{c|}{w/o DepthRB \& Gate} & -0.18 & 198.83\\
            \midrule
            \multirow{1}{*}{\textbf{LLIC-STF w/o CTB}}                   \\
            \multicolumn{1}{c|}{$K$ = \{5, 5, 5, 5\}}                   & +0.05     & 216.28                \\
            \multicolumn{1}{c|}{{$K$ = \{7, 7, 7, 7\}}}                   & -2.46     & 217.62                \\
            \multicolumn{1}{c|}{{$K$ = \{9, 9, 9, 9\}}} & -3.38 & 219.29\\
            \multicolumn{1}{c|}{{$K$ = \{11, 11, 9, 9\}}} & -4.19 & 221.29 \\
            \multicolumn{1}{c|}{{$K$ = \{11, 11, 11, 11\}}} & -4.42 & 221.62\\
            \midrule
            \multirow{1}{*}{\textbf{LLIC-STF}}\\
            \multicolumn{1}{c|}{256 $\times$ 256 patch}  &  -4.64  & 325.00 \\
            \multicolumn{1}{c|}{512 $\times$ 512 patch}  &  -7.85  & 325.00 \\
            \midrule
            \multicolumn{1}{c|}{VTM-17.0 Intra} & 0.00 & -- \\
            \bottomrule
            \end{tabular}
            \vspace*{-1\baselineskip}
            \end{table}
            \begin{table*}[h]   
                \centering
                \footnotesize
                \setlength{\tabcolsep}{1.6mm}{
                \caption{{Influence of Kernel Size on Various Resolutions.}}
                \label{tab:ks} 
                \begin{tabular}{@{}cccccccccccccc@{}}
                \toprule
                \multicolumn{1}{c|}{\multirow{2}{*}{{Kernel Size}}}                            &  \multicolumn{4}{c}{{BD-Rate (\%) w.r.t. {VTM-17.0 Intra}}}\\
                \multicolumn{1}{c|}{}                                                    & \multicolumn{1}{c}{{Kodak ($768\times 512$)~\cite{kodak}}}    & \multicolumn{1}{c}{{Tecnick ($1200\times 1200$)~\cite{tecnick2014TESTIMAGES}}}  & \multicolumn{1}{c}{{CLIC Pro Valid ($2048\times 1370$)~\cite{CLIC2020}}}  & \multicolumn{1}{c}{{JPEGAI Test ($3680\times 2456$)~\cite{jpegai}}}  \\\midrule
                \multicolumn{1}{c|}{{$K=\{5,5,5,5\}$}}                                        & {$+0.05$}       & {$-1.17$} & {$+1.72$}       & {$+1.13$}       \\\midrule
                \multicolumn{1}{c|}{{$K=\{7,7,7,7\}$}}                                        & {$-2.46$}       & {$-4.27$} & {$-1.63$}       & {$-2.91$}         \\\midrule
                \multicolumn{1}{c|}{{$K=\{9,9,9,9\}$}}                & {$-3.38$}      & {$-6.89$}   & {$-3.47$} & {$-5.12$}\\\midrule
                \multicolumn{1}{c|}{{$K=\{11, 11, 9, 9\}$}}                & {$-4.19$}      & {$-8.17$}   & {$-4.33$} & {$-7.76$} \\\bottomrule
                \end{tabular}}
              \end{table*}
\subsubsection{Qualitative Results}\label{sec:exp:perf:qual}
Fig.~\ref{fig:vis} illustrates the example of reconstructed Kodim07 
of our proposed models, WACNN, STF~\cite{zou2022the}, 
Xie'21~\cite{xie2021enhanced}, and VTM-17.0 Intra. 
The PSNR value of the image reconstructed by our proposed models is 
$1$ dB higher than that of the image reconstructed by VTM-17.0 Intra.
 Compared with STF, {VTM-17.0 Intra} and other codecs, 
 our proposed models can retain more details. 
 Images reconstructed by our proposed LLIC have higher subjective quality.
\subsection{Computational Complexity}\label{sec:exp:perf:eff}
GPU Memory Consumption-Image Resolution, Forward MACs-Image Resolution,
Encoding Time-Image Resolution, and Decoding Time-Image Resolution curves
are presented in Fig.~\ref{fig:complex}.
We compare our proposed LLIC-STF, LLIC-ELIC, and LLIC-TCM with their baseline models
STF~\cite{zou2022the}, ELIC~\cite{he2022elic}, and LIC-TCM Large~\cite{liu2023learned} and 
recent learned image compression models~\cite{cheng2020learned,qian2020learning,xie2021enhanced,qian2022entroformer,wang2022neural}.
Overall, our proposed transform coding method enhances the model performance while
exhibiting modest GPU memory, forward inference MACs, and fast encoding and decoding speeds.
\subsubsection{Testing GPU Memory Consumption Comparison}
Compared with existing recent learned image compression models Xie'21~\cite{xie2021enhanced},
Entroformer~\cite{qian2022entroformer}, and NeuralSyntax~\cite{wang2003multiscale},
our proposed LLIC transform coding consumes much less GPU memory.
When compressing $3072\times 3072$ images, the peak GPU memory of our
proposed LLIC models is {approximately} $10$ GB, while the GPU memory consumption of 
Xie'21~\cite{xie2021enhanced} is over $20$ GB.
Compared to the baseline models STF~\cite{zou2022the} and ELIC~\cite{he2022elic}, our proposed
LLIC-STF and LLIC-ELIC do not add much GPU memory overhead when compressing low-resolution images. 
The GPU memory consumptions of STF,
ELIC, LLIC-STF, and LLIC-ELIC are very similar for low-resolution images.
It should be emphasized that our LLIC-STF and LLIC-ELIC have great performance
improvements over STF~\cite{zou2022the} and ELIC~\cite{he2022elic},
which illustrates the superiority of our approach.
Compared to the baseline model LIC-TCM Large~\cite{liu2023learned}, our LLIC-TCM
significantly reduces the GPU memory consumption with better rate-distortion performance.
When compressing and decompressing $2048\times 2048$ images, the GPU memory consumption 
of our proposed LLIC-TCM is only $\frac{1}{3}$ of the GPU memory consumption of LIC-TCM Large.
When compressing $3072\times 3072$ images, the GPU memory consumption of our proposed LLIC-TCM 
is only $\frac{1}{4}$ of the GPU memory consumption of LIC-TCM Large.
The curve of the GPU memory consumed by our LLIC-TCM as the
resolution grows is much flatter.
The GPU memory overhead is a very important measure of model complexity because, once the amount
of GPU memory required for encoding or decoding exceeds the GPU memory capacity of the machine,
it is no {longer possible} to encode and decode images.
The lower GPU memory consumption indicates that our proposed LLIC-STF, LLIC-ELIC, and LLIC-TCM have lower complexity and 
are highly suitable for high-resolution image compression.
\subsubsection{Forward Inference MACs Comparison}
The forward inference MACs of the learned image compression models are computed through the fvcore, DeepSpeed, and Ptflops {libraries}.
Because the flops evaluation libraries are not very accurate, when evaluating one model,
we compute the output values of fvcore, Deepspeed, and Ptflops, and we take the average of the two closest values as the final result.
Compared with the recent learned image compression models Cheng'20~\cite{cheng2020learned}, Xie'21~\cite{xie2021enhanced},
Entroformer~\cite{qian2022entroformer}, and NeuralSyntax~\cite{wang2022neural}, our proposed LLIC-STF,
LLIC-ELIC and LLIC-TCM {demonstrate remarkable reductions in forward inference MACs}.
{Notably, the forward MACs required for our LLIC models to compress images of dimensions $2048 \times 2048$ are approximately one-fourth of those reported for NeuralSyntax.}
Compared with the baseline models STF~\cite{zou2022the} and ELIC~\cite{he2022elic}, our
LLIC-STF and LLIC-ELIC have slightly higher forward MACs. 
Considering the performance gains of our proposed LLIC-STF and LLIC-ELIC over STF~\cite{zou2022the}, and ELIC~\cite{he2022elic},
it is worthwhile to increase the forward MACs a bit.
In addition, the forward MACs of our proposed LLIC models are approximately $74\%$ of the forward MACs of Cheng'20~\cite{cheng2020learned}
when compressing $2048\times 2048$ images.
Compared to the baseline model LIC-TCM Large, our proposed LLIC-TCM significantly reduces the forward MACs.
The forward MACs of our proposed LLIC-TCM is only $45\%$ of the forward MACs of LIC-TCM Large when compressing $2048\times 2048$ images 
and compressing $3072\times 3072$ images.
Compared with the mixed CNN-Transformer-based transform coding of LIC-TCM Large,
our proposed large receptive field transform coding is more powerful and light-weight.
\subsubsection{Encoding and Decoding Time Comparison}
Because our LLIC-STF, LLIC-ELIC, and LLIC-TCM employ parallel entropy models, 
they encode and decode much faster than Cheng'20~\cite{cheng2020learned},
Xie'21~\cite{xie2021enhanced}, Qian'21~\cite{qian2020learning}, Entroformer~\cite{qian2022entroformer}
and NeuralSyntax~\cite{wang2022neural}. Specifically, when compressing a $768\times 768$
image, the encoding times of Cheng'20 and Xie'21 exceed $3$s, the encoding times
of Qian'21 and Entroformer exceed $40$s, the decoding times of Cheng'20 and Xie'21
exceed $7$s, and the decoding times of Qian'21 and Entroformer exceed $50$s; meanwhile, the 
encoding and decoding times of our proposed LLIC models are approximately $0.06\sim 0.1$s and $0.1\sim0.12$s, respectively.
Compared to the baseline model STF~\cite{zou2022the}, our proposed LLIC-STF is approximately as fast as STF~\cite{zou2022the}.
Compared to the baseline model ELIC~\cite{he2022elic}, our proposed LLIC-ELIC is approximately as
fast as ELIC~\cite{he2022elic} on low-resolution images.
Our proposed LLIC-ELIC requires $20\%$ more time
and requires $7\%$ to encode a $2048\times 2048$ image.
The added time to encode and decode the high-resolution images 
is worthwhile considering the performance improvement brought about
by our proposed transform coding. Compared with LIC-TCM Large,
our LLIC-TCM encodes and decodes much faster than {the} baseline model LIC-TCM Large.
Overall, our proposed transform coding enhances model performance while achieving
fast encoding and decoding speed.
\subsection{Ablation Studies}\label{sec:exp:ablation}
\subsubsection{Settings}
Ablation studies are conducted on LLIC-STF.
When conducting ablation experiments, we train each model for $1.7$M steps from scratch.
The batch size is set to $16$.
We adopt the training strategy in section~\ref{sec:exp:imple:train}.
The results of ablation studies {are} in Table~\ref{tab:ablation}.
In Table~\ref{tab:ablation}, “w/o Gate” means that the gate block is replaced by {an} FFN~\cite{vas2017attention}.
\subsubsection{Influence of Self-Conditioned Spatial Transform}\label{sec:exp:ablation:ks}
Although it is possible to increase the theoretical receptive
field by continuously stacking network layers,
it is likely that a small effective receptive field~\cite{luo2016understanding} will be obtained.
Following Zhu \textit{et al}~\cite{zhu2022transformerbased}, 
Effective Receptive Field (ERF)~\cite{luo2016understanding,ding2022scaling} is employed to
evaluate the influence of the proposed self-conditioned spatial transform, which
utilizes large depth-wise convolutions. 
The ERF is visualized as the absolute gradient of the center pixel in the latent ($\mathrm{d}\boldsymbol{y}/\mathrm{d}\boldsymbol{x}$)
with respect to the input image.
The ERFs of our baseline models
STF~\cite{zou2022the}, ELIC~\cite{he2022elic}, and LIC-TCM Large~\cite{liu2023learned}
, our proposed LLIC-STF, LLIC-ELIC and LLIC-TCM are visualized in Fig.~\ref{fig:erf}.
Baseline model STF employs swin-transformer-based transform coding,
ELIC employs CNN-Attention-based transform coding, and the recent
LIC-TCM employs mixed CNN-Transformer-based transform coding. Compared 
with these different types of transform coding methods,
it is obvious that our proposed LLIC-STF, LLIC-ELIC, and LLIC-TCM achieve 
much larger ERFs than their baselines.
Larger ERFs indicate that our proposed large receptive field transform coding is 
able to remove more redundancy during the analysis transform, which
makes our proposed models perform better than our baseline models.\par
{To further demonstrate the advantages of large receptive fields, we compute the
    cosine similarity between the center latent and other latents in ${\boldsymbol{y}}$.
    The cosine similarity computation is performed on a $17\times 17$ window obtained by center-cropping the latent representation. 
    We compute cosine similarity on the Kodak dataset. The cosine similarities are averaged on $24$ images from the Kodak dataset.
    The absolute cosine similarity map is visualized in Figure~\ref{fig:cs}. 
    According to Figure~\ref{fig:cs}, our proposed method can reduce a large range of redundancies.
    For example, compared to the baseline methods STF, ELIC, and LIC-TCM, 
    the similarity values of the $16$-th rows of the similarity maps of the proposed method are lower.
    The lower similarity among distant latents can be attributed to larger effective receptive fields.
    A lower similarity indicates that a larger receptive field can reduce more spatial redundancy.}\par
To further investigate the contribution of {the} proposed spatial transform block,
the spatial transform block is removed or replaced in our ablation studies.
Specifically, removing the STB leads to significant performance degradation.
If the STB is replaced by CTB, the rate-distortion performance is still
not as good as the performance of the model utilizing STBs and CTBs.
This performance degradation indicates the necessity of employing the proposed
spatial transform blocks for a more compact latent representation to enhance
the rate-distortion performance.\par
In our models and baseline models, the analysis transform $g_a$ and 
synthesis transform $g_s$ involves four stages. The kernel
size of each stage is denoted as $K=\{k_1, k_2, k_3, k_4\}$.
To evaluate the contribution of large kernel, we set the
kernel size $K=\{5,5,5,5\}$, $K=\{7,7,7,7\}$, $K=\{9,9,9,9\}$, $K=\{11,11,9,9\}$, 
and $K=\{11,11,11,11\}$.
The rate-distortion performances when utilizing various kernel sizes are
shown in Table~\ref{tab:ablation}.
{Clearly, increasing the size of the convolutional
kernel keeps improving the rate-distortion performance of the model,
However, the gains decrease.}
Increasing the kernel size $K$ from $\{5,5,5,5\}$ to $\{7,7,7,7\}$ leads to
the largest performance enhancement.
The differences in the performance when utilizing $\{11,11,9,9\}$ kernels and 
when utilizing $\{11,11,11,11\}$ kernels are quite negligible.
Therefore, in our proposed LLIC models, $K=\{11,11,9,9\}$ is employed.\par
{
To {analyze} the relationship between the kernel size and compression performance with different resolutions, we conduct
ablation studies on Kodak, Tecnick~\cite{tecnick2014TESTIMAGES}, CLIC Pro Valid~\cite{CLIC2020}, and JPEGAI Test~\cite{jpegai}.
The results are presented in Table~\ref{tab:ks}.
As the resolution of the input image increases, the gap between the {BD-Rate} values of different kernels gradually increases.
For example, when compressing $512\times 768$ images (Kodak), the difference between $K=\{7,7,7,7\}$ and $K=\{5,5,5,5\}$ is 
$-2.51\%$; when compressing $1200\times 1200$ (Tecnick) images, $2048\times 1370$ (CLIC Pro Valid) images, and $3680\times 2456$
(JPEGAI Test) images, the differences are $-3.10\%, -3.35\%, -4.04\%$, respectively.
The increased BD-Rate gap indicates that large kernel sizes is beneficial for high-resolution image coding.
Compared with low-resolution images, high-resolution images contain more \textit{spatial redundancy}.
For high-resolution images, pixels in a larger area are correlated with each other.
In this case, the receptive field of the image compression model is crucial.
If the receptive field is small, redundancy beyond its receptive field will be difficult to remove, 
especially when compressing high-resolution images.
Therefore, we employ large depth-wise kernels to overcome the drawbacks of the previous methods
while maintaining modest complexity. The large kernel leads to better performance when compressing high-resolution images.}\par
Context adaptability also plays an important role in boosting the rate-distortion
performance of learned image compression models.
When the kernel weights are independent of the input feature, the
rate-distortion becomes worse. The rate-distortion performance loss is approximately $1.6\%$.
The performance degradations demonstrate the effectiveness of the proposed 
self-conditioned weight generation.
\subsubsection{Influence of Self-Conditioned Channel Transform}\label{sec:exp:ablation:crb}
The channel transform block is proposed for self-conditioned adaptive channel adjustment.
In ablation studies, channel transform blocks are removed or replaced.
Specifically, removing CTBs leads to significant performance degradation, and 
the rate-distortion performance of replacing CTBs with STBs is still not as good as
the performance of the model utilizing CTBs and STBs.
The performance degradation in ablation studies demonstrates the
effectiveness and necessity of employing channel transform blocks in transform coding.
\subsubsection{Influence of DepthRB and Gate Mechanism}\label{sec:exp:ablation:rb}
We use LLIC-STF without CTBs to evaluate the effectiveness of the
proposed DepthRB for nonlinear embedding.
Compared with linear embedding, which employs a linear layer, 
our proposed DepthRB for non-linear embedding further enhances the rate-distortion
performance. Non-linear embedding is more flexible than linear embedding.
We also evaluate the effectiveness of the proposed gate block.
Replacing the gate blocks with vanilla FFNs increases the complexity of the model but
leads to performance degradation, which demonstrates the superiority of the
proposed gate block.
\subsubsection{Influence of Large Patches for Training}
The $256\times 256$ patches are insufficient for training.
To fully exploit large kernels, the large training strategy is employed.
{Large patch training results in a performance increment of approximately $3\%$,
thereby substantiating the efficacy of the large patch training strategy.}
\section{Conclusion}
In this paper, we propose large receptive-field transform coding with
self-conditioned adaptability for learned image compression,
which effectively captures more spatial correlations. To reduce channel-wise redundancy,
we propose the self-conditioned channel transform
to adjust the weight of each channel.
To evaluate our proposed transform method, we align the entropy model with 
existing advanced non-linear transform coding techniques and obtain the models
LLIC-STF, LLIC-ELIC, and LLIC-TCM.
Extensive experiments demonstrate the superiority of our proposed
large receptive field learning with self-conditioned adaptability.
Our LLIC-STF, LLIC-ELIC, and LLIC-TCM achieve state-of-the-art performance
and they reduce the {BD-Rate} by $9.49\%, 9.47\%, 10.94\%$ on Kodak
over VTM-17.0 Intra, respectively.
To further enhance the performance, it is promising to integrate with
more advanced entropy models~\cite{jiang2022mlic,jiang2023mlic++}.
{However, there are several limitations to be addressed.
First, the complexities of the proposed LLIC models are not low enough, which means that {they cannot be directly employed on mobile devices}. Second, the learned image compression models are trained on natural images, which means that the generalization ability on {out-of-distribution} images (e.g., screen content) may be limited.
The decoding complexity of LLIC models can be further reduced by employing asymmetric encoder-decoder structure~\cite{yang2023asymmetrically}, where
the encoder could be heavy while the decoder is light. To enhance the generalization ability of learned image compression
models, we suggest fine-tuning the encoder or latent representation for a specific input, which will increase the encoding
time, but the decoding complexity will still be low. In addition, fine-tuning the encoder or latent representation will also improve the
performance on natural images. We will investigate these techniques in the future.}

\bibliographystyle{IEEEtran}
\bibliography{reference}


\begin{IEEEbiography}[{\includegraphics[width=1in,height=1.25in,clip,keepaspectratio]{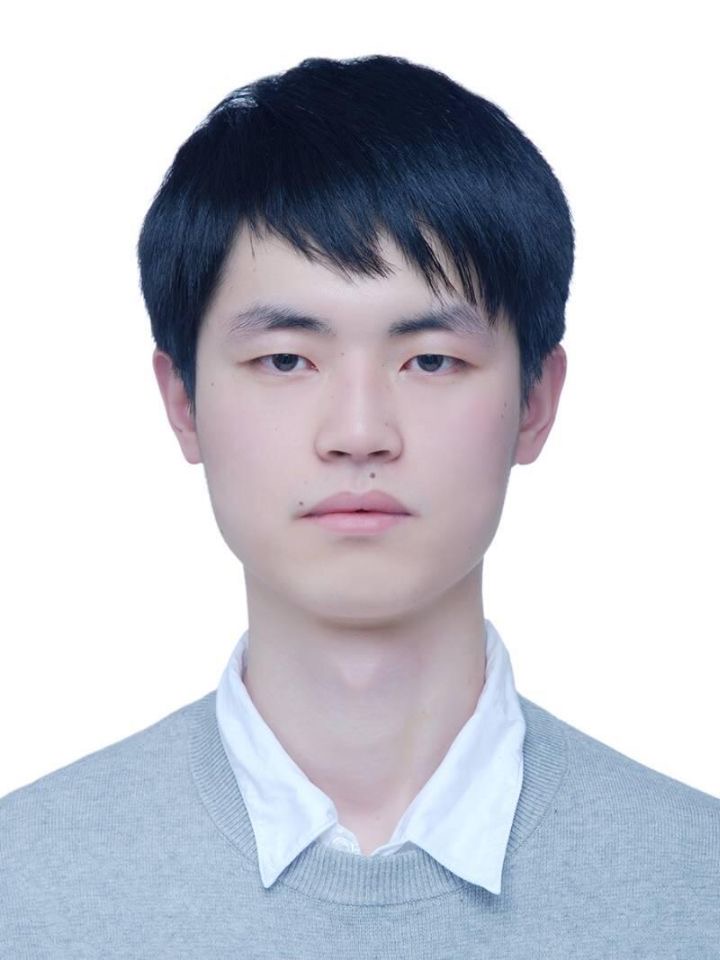}}]{Wei Jiang} received the B.S. degree in Vehicle Engineering from Chongqing University, China, in 2021. He is currently pursuing the Ph.D. degree in computer application technology at Peking University. His research interests include learned image / video coding and implicit neural representation.
\end{IEEEbiography}
\begin{IEEEbiography}
[{\includegraphics[width=1in,height=1.25in,clip,keepaspectratio]{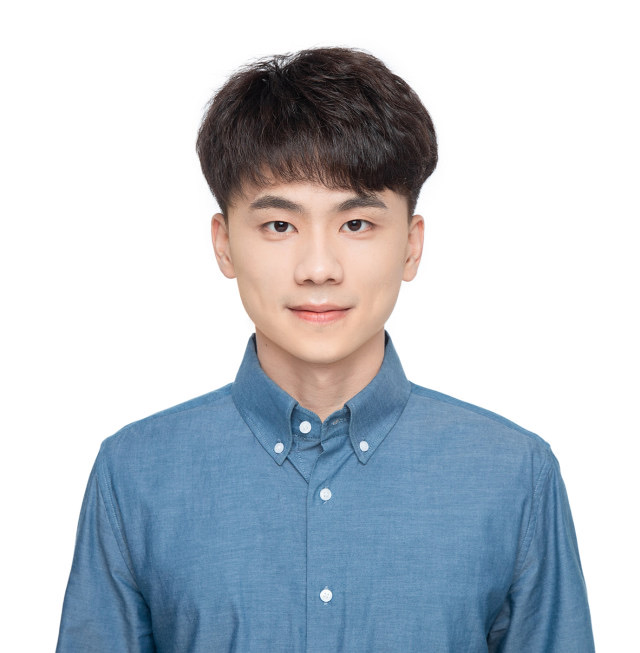}}]{Peirong Ning} received his Master's degree in Computer Science from Peking University in 2023. He is currently working at Xiaohongshu Company, Ltd., where he focuses on learning-based codecs and the Hybrid Video Coding framework.
\end{IEEEbiography}
\begin{IEEEbiography}
[{\includegraphics[width=1in,height=1.25in,clip,keepaspectratio]{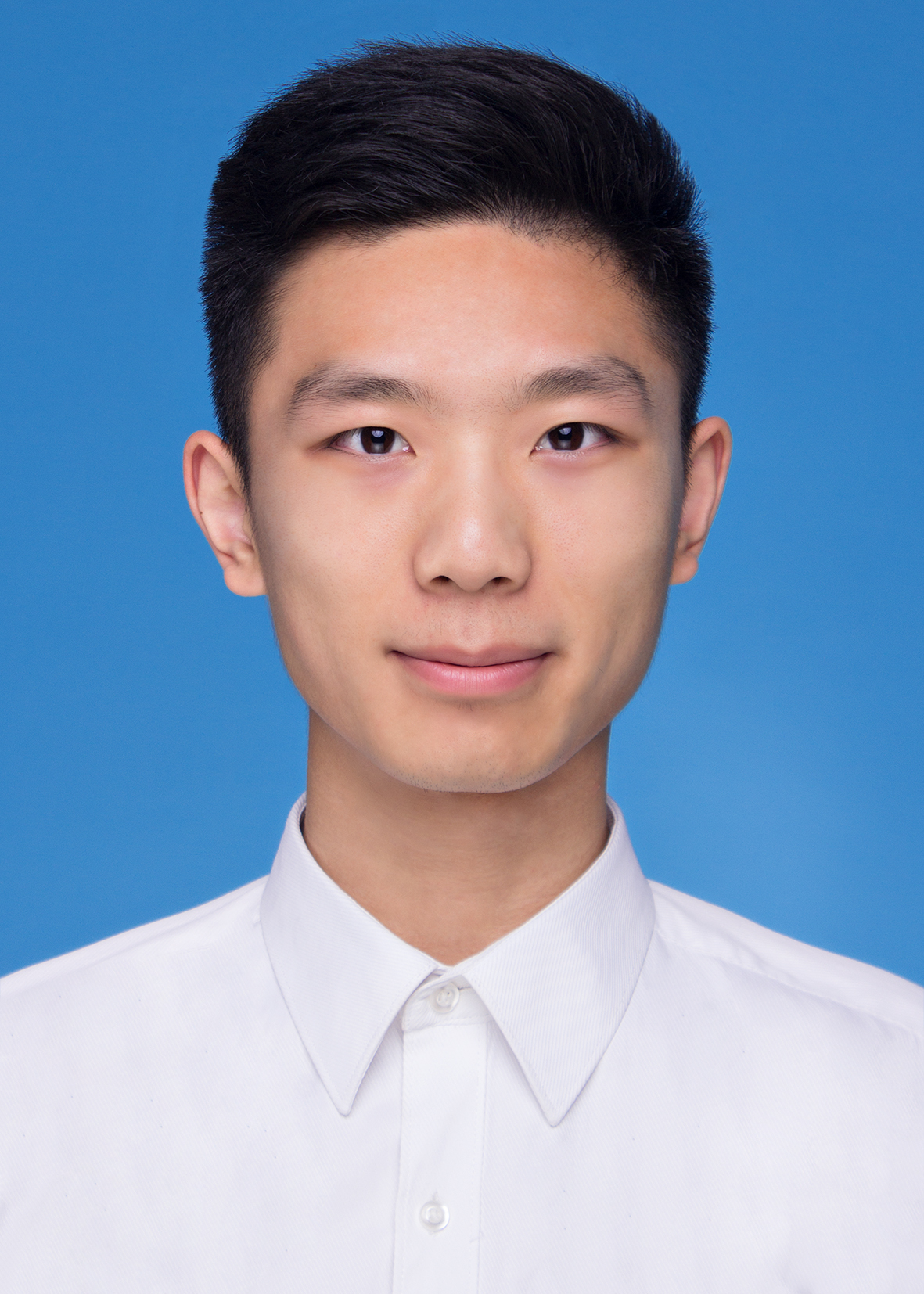}}]{Jiayu Yang} received the B.E. degree in electronic information engineering from Hefei University of Technology, China, in 2019. He is currently pursuing the Ph.D. degree in the School of Electronic and Computer Engineering at Peking University, China. His research interests include video coding and immersive media coding.
\end{IEEEbiography}
\begin{IEEEbiography}[{\includegraphics[width=1in,height=1.25in,clip,keepaspectratio]{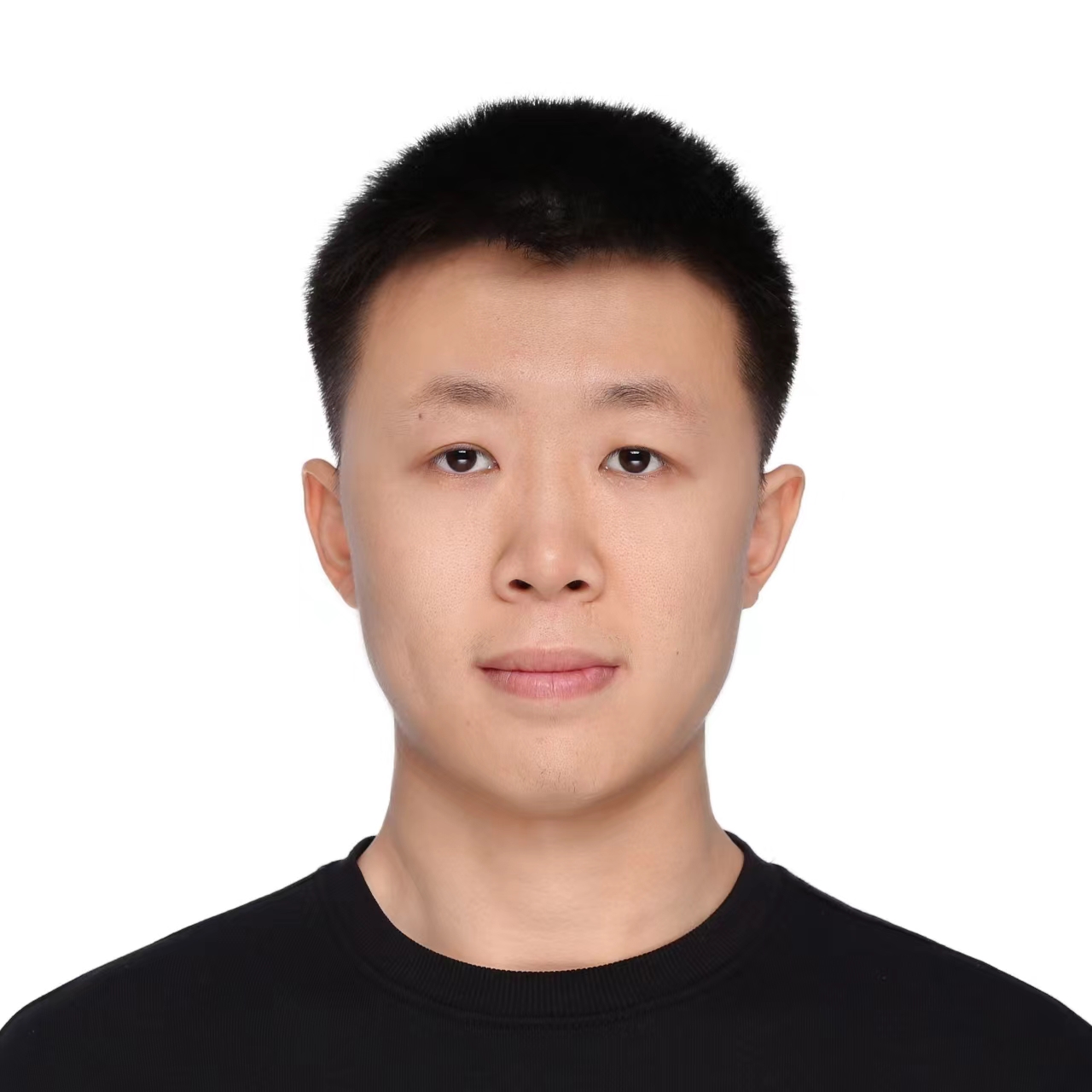}}]{Yongqi Zhai} received the B.S. degree in electronic information engineering from Jilin University, China, in 2020. He is currently pursuing the Ph.D. degree in computer application technology at Peking University. His research interests include traditional and learning-based image/video coding.
\end{IEEEbiography}
\begin{IEEEbiography}[{\includegraphics[width=1in,height=1.25in,clip,keepaspectratio]{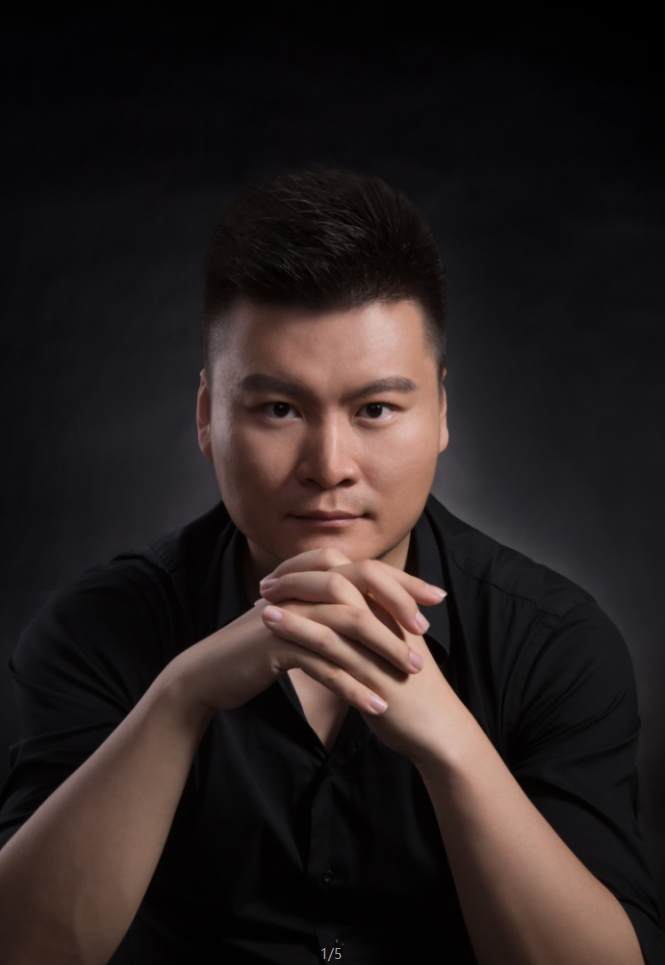}}]{Feng Gao} (Member, IEEE) received his B.S. degree in Computer Science from University College London in 2007, and Ph.D. degree in Computer Science from Peking University in 2018. He was a post-doctoral research fellow at the Future Laboratory,Tsinghua University, from 2018 to 2020. He joins Peking University as Assistant Professor since 2020. His research interesting is working on the intersection of Computer Science and Art, including but not limit on artificial intelligence and painting art, deep learning and painting robot, etc.
\end{IEEEbiography}
\begin{IEEEbiography}[{\includegraphics[width=1in,height=1.25in,clip,keepaspectratio]{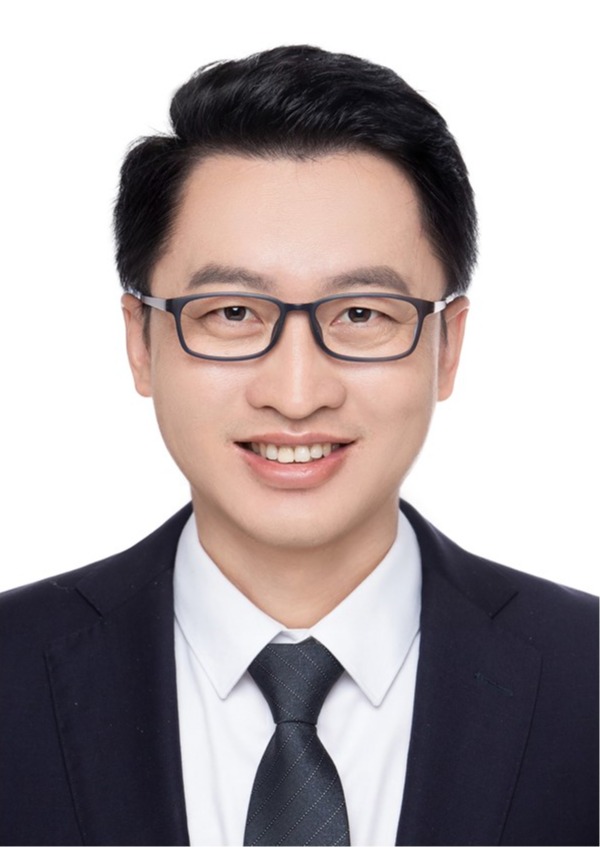}}]{Ronggang Wang} (Member, IEEE) received the Ph.D. degree from the Institute of Computing Technology, Chinese Academy of Sciences, in 2006. He is currently a Professor with the School of Electronic and Computer Engineering, Peking University Shenzhen Graduate School. His research interests include immersive video coding and processing. He has made over 100 technical contributions to ISO/IEC MPEG, IEEE 1857 and China AVS. He has authored more than 150 papers and held more than 100 patents. He has been serving as the IEEE 1857.9 Immersive video coding standard sub-group Chair and China AVS virtual reality sub-group Chair since 2016.
\end{IEEEbiography}

\vfill

\end{document}